\documentclass[lettersize,journal]{IEEEtran}
\usepackage{amsmath,amsfonts}
\usepackage{algorithmic}
\usepackage{array}

\usepackage{textcomp}
\usepackage{stfloats}
\usepackage{url}
\usepackage{verbatim}
\usepackage{graphicx}
\def\BibTeX{{\rm B\kern-.05em{\sc i\kern-.025em b}\kern-.08em
    T\kern-.1667em\lower.7ex\hbox{E}\kern-.125emX}}
\usepackage{balance}

\usepackage{xcolor}
\usepackage{multirow}
\usepackage{makecell}  
\usepackage{dsfont}
\usepackage{float} 
\usepackage{booktabs}
\usepackage{threeparttable}

\usepackage{subfigure}
\usepackage{hyperref}

\usepackage{amssymb}

\begin{document}
\title{MARG: MAstering Risky Gap Terrains for Legged Robots with Elevation Mapping}

\author{Yinzhao Dong$^{*}$, Ji Ma$^{*}$, Liu Zhao, Wanyue Li, Peng Lu$^{\dagger}$

\thanks{Manuscript created 19 November 2024; Revised 31 July 2025; Accepted 20 September 2025. This work was supported by the General Research Fund under Grant 17204222, and in part by the Seed Fund for Collaborative Research and General Funding Scheme-HKU-TCL Joint Research Center for Artificial Intelligence. This paper was recommended for publication by Editor Jens Kober and Editor-in-Chief Wolfram Burgard upon evaluation of the reviewers' comments. (Corresponding author: Peng Lu).}

\thanks{The authors are with the Adaptive Robotic Controls Lab (ArcLab), Department of Mechanical Engineering, The University of Hong Kong, Hong Kong SAR, China (E-mail: \url{dongyz@connect.hku.hk}, \url{maji@connect.hku.hk}, \url{zhaol@connect.hku.hk}, \url{liwy1024@connect.hku.hk}, \url{lupeng@hku.hk}.}

\thanks{$^{*}$ Equal Contribution.} 
\thanks{The supplementary video is available at \protect\url{https://youtu.be/NOVmjvWUM8Y}}
\thanks{Digital Object Identifier (DOI): ras.tro.25-0713.aa6b0eec.}

}

\markboth{IEEE Transactions on Robotics. Preprint Version. Accepted September, 2025}
{Dong \MakeLowercase{\textit{et al.}}: MARG: MAstering Risky Gap Terrains for Legged Robots with Elevation Mapping} 

\maketitle

\begin{abstract}
Deep Reinforcement Learning (DRL) controllers for quadrupedal locomotion have demonstrated impressive performance on challenging terrains, allowing robots to execute complex skills such as climbing, running, and jumping. However, existing blind locomotion controllers often struggle to ensure safety
and efficient traversal through risky gap terrains, which are typically highly complex, requiring robots to perceive terrain information and select appropriate footholds during locomotion accurately. Meanwhile, existing perception-based controllers still present several practical limitations, including a complex multi-sensor deployment system and expensive computing resource requirements. This paper proposes a DRL controller named MAstering Risky Gap Terrains (MARG), which integrates terrain maps and proprioception to dynamically adjust the action and enhance the robot's stability in these tasks. During the training phase, our controller accelerates policy optimization by selectively incorporating privileged information (e.g., center of mass, friction coefficients) that are available in simulation but unmeasurable directly in real-world deployments due to sensor limitations. We also designed three foot-related rewards to encourage the robot to explore safe footholds. More importantly, a terrain map generation (TMG) model is proposed to reduce the drift existing in mapping and provide accurate terrain maps using only one LiDAR, providing a foundation for zero-shot transfer of the learned policy. The experimental results indicate that MARG maintains stability in various risky terrain tasks.
\end{abstract}


\section{Introduction}

\begin{figure}[h]
\centering\includegraphics[width=0.46\textwidth,  trim=10 10 10 10,clip]{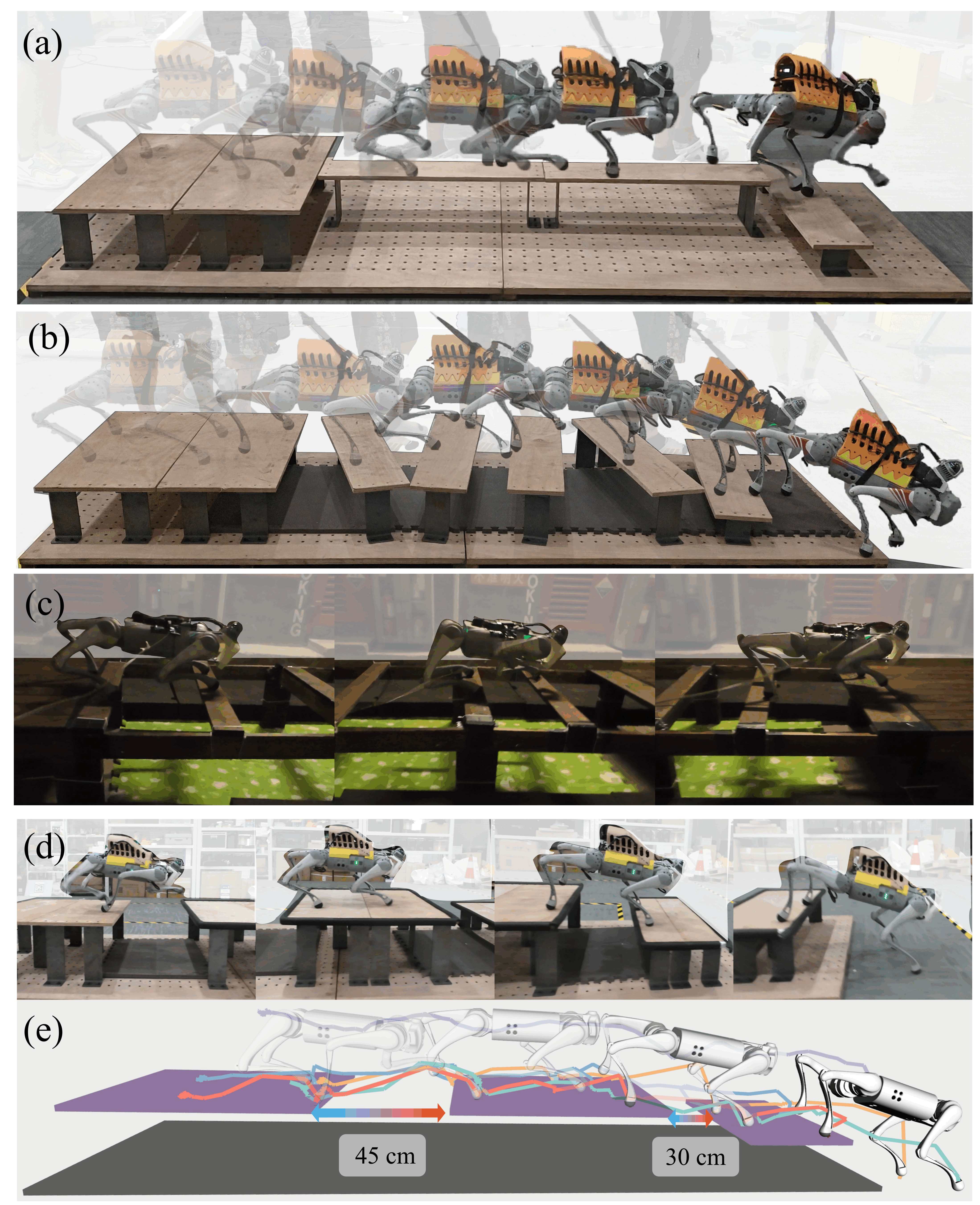}
   \caption{Experiment of Unitree Go1 and Go2 on risky gap terrains, including (a) Single plank bridge, with the narrowest traversable width being 18 cm, validates the center-of-gravity control under narrow support. (b-c) Balance beams, where the narrowest beam width is 9 cm, to test the robot's stability in response to height variations, inclination changes, and edge perception. (d-e) Large gaps to demonstrate the capability to traverse gaps of varying widths (up to 65 cm in the real-world experiment).}
   \label{pic_main_hardware}
\end{figure} 

\IEEEPARstart{L}egged robots have significantly advanced locomotion capabilities, demonstrating impressive skills across various movement modes, such as climbing stairs \cite{Shamel2023,luo2024moral}, descending ramps \cite{Abdalla2023}, high-speed running \cite{ji2022concurrent}, parkour \cite{cheng2024extreme}, bipedal locomotion \cite{xiao_learning_2025}, and backflipping \cite{kim2024stage}. These abilities enable robots to perform well in continuous and highly challenging terrains, including rugged mountain paths, narrow passages, stairwells, slippery or unstable surfaces, etc. However, existing blind locomotion controllers often struggle to overcome risky gap terrains due to shortcomings in ensuring the safety and balance of quadruped robots.

Risky gap terrains exhibit numerous complex characteristics, imposing nearly stringent demands on robots regarding footholds and balance capabilities during locomotion. As shown in Fig. \ref{pic_main_hardware} (a), robots must not only strive to maintain the stability of their center of gravity on a narrow single-plank bridge but also respond in real time to potential lateral disturbances. Once a robot makes errors during locomotion, such as slipping or shifting its center of gravity, it may quickly step on the air or lose stability, leading to a fall and potentially causing severe damage to the robot. When traversing balance beams, the quadruped robot must not only accurately perceive terrain information such as height variations, gap width, and edges, but also select appropriate landing footholds and timing for exertion based on its locomotion capabilities and current state to avoid missteps, as shown in Figs. \ref{pic_main_hardware} (b-c). 

The majority of existing quadrupedal locomotion controllers are blind, which means that they do not utilize perception sensors like cameras and LiDARs \cite{hwangbo2019learning, lee2020learning, nahrendra2023dreamwaq, Margolis2023, Luo2023}. It is nearly impossible for these controllers to traverse risky terrains as shown in Fig.~\ref{pic_main_hardware}. Recently, perception sensors have been used to obtain an elevation map of the environment \cite{miki2022learning, Shamel2023}. However, they do not take risky terrains into consideration. Only a few studies consider risky terrains, and they either rely on multiple sensors \cite{jenelten2024dtc}, which significantly increases the complexity of hardware deployment, or use motion capture systems to obtain prior information about the terrain \cite{zhang2023learning}. In this paper, we only use one sensor to construct a robot-centered map and do not rely on motion capture systems.

\subsection{Model-based Legged Locomotion} 
Existing model-based controllers rely on precise modeling of robots to calculate the optimal joint torques or footholds required for locomotion. For example, Singh et al. \cite{singh2023analytical} compute second-order derivatives of rigid-body inverse and forward dynamics, achieving significant speed-ups over automatic differentiation in optimization-driven robot control. The CAFE-MPC framework \cite{li2024cafe} employs a cascaded-fidelity model predictive control scheme paired with a tuning-free whole-body controller, enabling quadruped robots to execute agile maneuvers without manual parameter tuning. Meduri et al. \cite{meduri2023biconmp} splits the nonlinear MPC problem into biconvex centroidal dynamics and full-body kinematics, enabling real-time generation of dynamic whole-body motions for legged robots. These models can generate accurate control commands, enabling the robot to achieve stable and efficient locomotion in an ideal simulation and simple terrains \cite{holmes2006dynamics}. However, uncertainty factors in real-world environments, such as terrain irregularity, changing friction, and external disturbances, present significant challenges to model-based methods. These factors are difficult to accurately incorporate into models, leading to potential mismatches between the model and the real world. Even slight discrepancies can cause robot locomotion failures, especially in risky gap terrains. 

To address these challenges, researchers \cite{grandia2023perceptive, Meduri2023} have attempted to simplify the dynamics model by utilizing Nonlinear Model Predictive Control (NMPC) to enhance the locomotion of robots in complex and dynamic environments. Yin et al. \cite{yin2023footholds} propose an optimization algorithm to improve the robot's locomotion performance by transforming the discrete terrain height map into a continuous cost map to adjust the footholds dynamically.  \cite{sombolestan2024adaptive} proposes a novel control system that integrates adaptive control into a force-based control system for legged robots, enabling them to dynamically locomotion on uneven terrains. However, the computational complexity and slow convergence rates limit the applicability of robots in dynamic environments.

In addition, studies \cite{fahmi2022vital, bellicoso2018advances} are also exploring the use of multiple sensors to enhance the accuracy and reliability of the model. Alongside the robot's inertial measurement unit (IMU), external perception devices such as depth cameras and LiDARs are employed to gather environmental information, including terrain width, height, and edge shape, and integrate this information into the dynamic model to assist robot control \cite{bjelonic2021whole}. The synchronization of sensor data, the design of fusion algorithms for different sensor inputs, and the computational burden of data processing will further adversely affect the real-time control performance of robots.

\subsection{Model-free Legged Locomotion} 
Model-free methods, such as deep reinforcement learning (DRL), have shown promise in enabling legged robots to adapt to complex terrains without relying on precise dynamic models. These methods focus on training robots to learn optimal policies through trial and error, allowing them to manage uncertainties and dynamic changes in their environment effectively \cite{sutton1998reinforcement}. The blind locomotion controllers \cite{luo2024moral, nahrendra2023dreamwaq} have shown impressive progress in enabling robots to traverse challenging continuous terrains. However, these controllers often struggle in risky terrains due to the absence of environmental perception.

Integrating data from other external sensors, such as depth cameras, motion capture, etc, into the DRL framework is an effective way to help robots comprehensively understand their surrounding environment. Pioneering works \cite{zhuang2023robot, cheng2024extreme, zhuang2024humanoid} utilize deep learning models, such as GRU \cite{dey2017gate} and LSTM \cite{graves2012long}, to process depth images and extract terrain features, including height, slope, and distribution. Robots can successfully perform high-difficulty parkour tasks by incorporating these terrain factors into their decision-making processes. Challenges such as lighting changes, occlusion issues, high dimensionality, and complexity of images \cite{dong2022towards} may lead to inaccurate terrain feature extraction or high computational complexity during the training process, ultimately affecting the real-time performance of the robot \cite{agarwal2023legged}. Meanwhile, \cite{shi2023terrain} and \cite{zhang2023learning} use motion capture and an offline map to derive the height map around the robot's feet,  which limits the practical applicability of this algorithm.

\subsection{Elevation Mapping for Legged Robots} 
To obtain more accurate terrain information in the real world, previous DRL controllers \cite{miki2022learning, hoeller2024anymal, jenelten2024dtc} utilize multiple depth cameras or LiDARs simultaneously for elevation mapping, which can significantly enhance the accuracy of terrain representation. However, this approach increases the complexity of hardware deployment, as it requires sophisticated processing capabilities to handle the data from multiple sensors. Additionally, existing localization technologies ~\cite{bledt2018cheetah, shan2020lio, zhang2014loam, xu2022fast} heavily rely on the pose estimation of floating bases within the global frame. Any inaccuracy in this estimation may lead to map drift, thereby affecting the movement of legged robots in risky terrains. Thus, designing safe and reliable controllers, developing efficient algorithms to simplify deployment processes, and obtaining precise terrain maps remain challenging in risky gaps tasks.

In summary, we propose a DRL controller for quadrupedal locomotion—MAstering Risky Gap (MARG)—which integrates terrain maps, privileged information, and proprioceptive into the policy to enhance the locomotion performance of quadrupedal robots in risky terrains. The key contributions of this work can be listed as follows:
\begin{itemize}
    \item We propose a safe and robust robot controller for locomotion, which can predict the body velocity and the contact state of feet on each step, significantly enhancing the robot's stability in risky gap terrains. 
    \item For risky tasks, we have designed three foot-related rewards: feet air time, feet stumble, and feet center, which promote the policy to explore safe footholds, enhancing the safety of movement.
    \item We propose a terrain map generation model that uses a single LiDAR to obtain the robot-centered height map. Our method minimizes drift compared to the traditional localization approaches while achieving zero-shot transfer capability and optimal computational efficiency.
    \item The MARG controller empowers quadruped robots to adeptly handle risky gap terrains in the real world, including 65 cm large gaps, 18 cm narrow single-plank bridges, and balance beams with varying sizes, heights, and inclinations.
\end{itemize}

\section{MARG Locomotion Controller}\label{methodology}
\begin{figure*}[h]
\centering\includegraphics[width=0.98\textwidth,trim=2 2 2 2,clip]{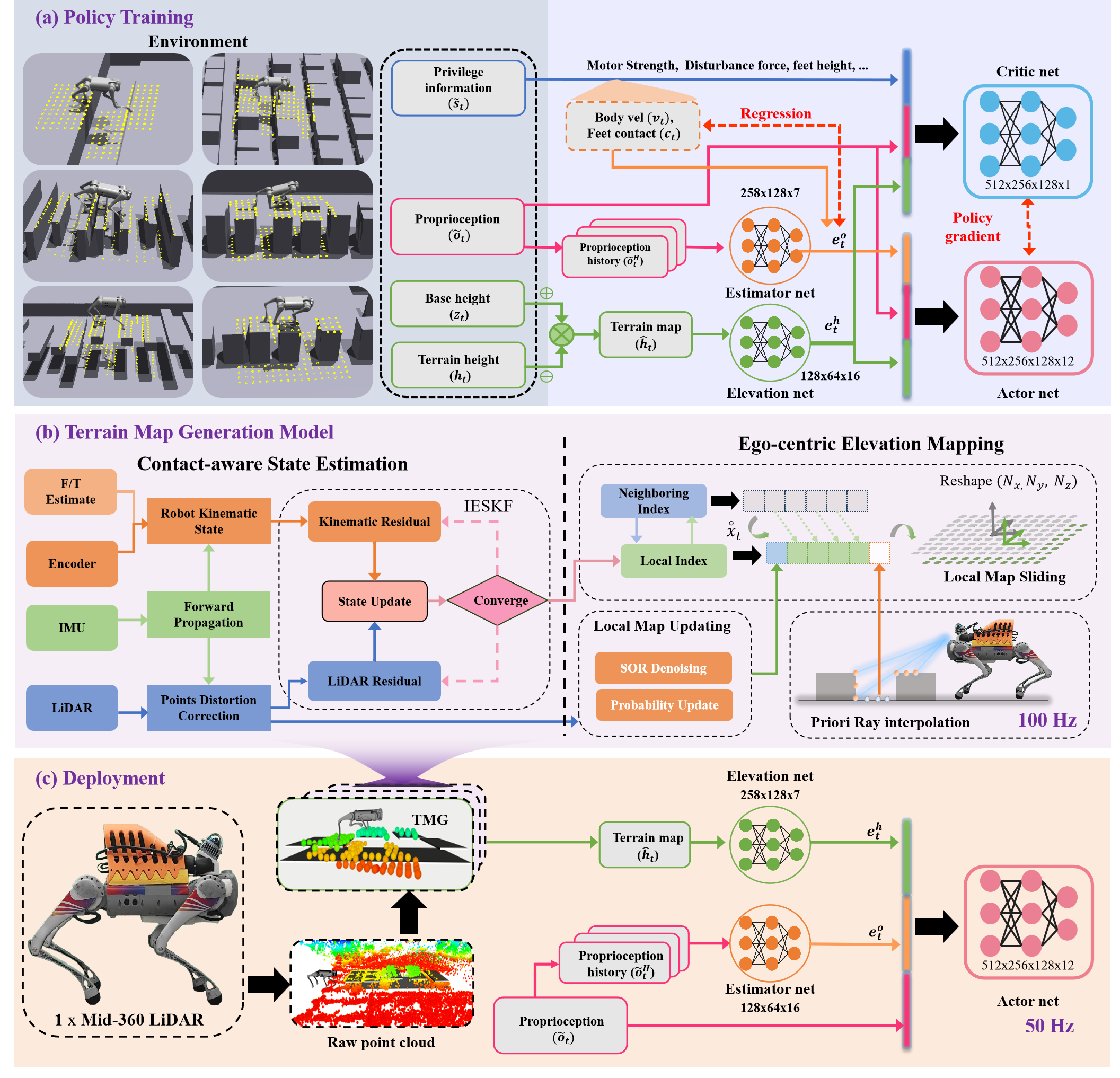}
   \caption{The framework of the proposed MARG, showing (a) Policy training module including actor, critic, estimator, and elevation nets, along with their data flow; (b) Terrain map generation module (TMG) including the processes of state estimation, mapping, and interpolation; (c) Deployment module indicating how the MARG is implemented in the real-world with a Mid-360 LiDAR inputs and network operations.}
   \label{pics_framework}
\end{figure*}
\subsection{Problem Formulation}\label{Formulation}
Terrain-aware legged locomotion task is a type of sequential decision-making problem under uncertainty. However, even with exteroceptive sensors, such as LiDARs and depth cameras, certain privileged information (e.g., the body velocity, the mass of each link, and the friction) still cannot be accurately obtained in real robots. Therefore, our problem can be formulated as an infinite-horizon Partially Observable Markov Decision Process (POMDP), denoted as a 7-tuple $\mathcal{M} = \{ \mathcal{S}, \mathcal{O}, \mathcal{A}, \mathcal{R}, \mathcal{P}, \Omega, \gamma\}$. Here, $\mathcal{S}$, $\mathcal{O}$, and $\mathcal{A}$ are the sets of states, observations, and actions, respectively. For each state $\boldsymbol{s}_t \in \mathcal{S}$, the learning agent interacts with the environment with an action  $\boldsymbol{a}_{t} \in  \mathcal{A}$ and receives a reward $\mathcal{R}(\boldsymbol{s}_t, \boldsymbol{a}_t)$, leading to the transition of the environment to the next state $\boldsymbol{s}_{t+1}$ with the probability $\mathcal{P}(\boldsymbol{s}_{t+1}|\boldsymbol{s}_t, \boldsymbol{a}_t)$. Meanwhile, the observation $\boldsymbol{o}_{t+1} \in \mathcal{O}$ depends on the new state $\boldsymbol{s}_{t+1}$ and the action 
$\boldsymbol{a}_t$ with a conditional probability $\Omega (\boldsymbol{o}_{t+1}|\boldsymbol{s}_{t+1}, \boldsymbol{a}_t)$. The objective is to determine the optimal policy $\pi^{*}$ that maximizes the accumulated rewards.
\subsubsection{Observation Space}
The proprioceptive observations $\boldsymbol{\widetilde{o}}_t = \left( \boldsymbol{\omega}_t,\ \boldsymbol{g}_t,\ \boldsymbol{v}^{*}_{t},\ \boldsymbol{q}_{t},\ \boldsymbol{\dot{q}}_{t}, \ \boldsymbol{a}_{t-1} \right)$ in our task include 
body angular velocity $\boldsymbol{\omega}_t\in\mathbb{R}^{3}$, projected gravity $\boldsymbol{g}_t\in\mathbb{R}^{3}$, linear velocity command $\boldsymbol{v}^{*}_{t} \in \mathbb{R}^{3}$, joint angles $\boldsymbol{q}_{t}\in\mathbb{R}^{12}$, joint angular velocities $\boldsymbol{\dot{q}}_{t}\in\mathbb{R}^{12}$, and the action of the last step $\boldsymbol{a}_{t-1}\in\mathbb{R}^{12}$. Meanwhile, we define a temporal observations $\boldsymbol{\widetilde{o}}_t^H= \left[ \boldsymbol{\widetilde{o}}_t, \boldsymbol{\widetilde{o}}_{t-1}, ..., \boldsymbol{\widetilde{o}}_{t-H} \right]$ to store the proprioceptive observations over the past $H$ time step ($H=5$ in this task). In addition, we also collect the exteroception observations (i.e., the egocentric terrain map of the robot body $\boldsymbol{h}_{t} \in \mathbb{R}^{187}$) and the privileged state $\boldsymbol{\widetilde{s}}_t$ defined as:
\begin{equation}
  \boldsymbol{\widetilde{s}}_t = (\boldsymbol{v}_t, 
    \ \boldsymbol{c}_t, 
    \ \boldsymbol{m}, 
    \ \boldsymbol{\mu},
    \ \boldsymbol{\zeta}, 
    \ \boldsymbol{f}_t, 
    \ \boldsymbol{k}_{pd})
\end{equation}
where $\boldsymbol{v}_t \in \mathbb{R}^{3}$, $\boldsymbol{c}_t \in \mathbb{R}^{4}$, $\boldsymbol{m} \in \mathbb{R}^{4}$,
$\boldsymbol{\mu} \in \mathbb{R}^{1}$, $\boldsymbol{\zeta} \in \mathbb{R}^{2}$, and  $\boldsymbol{f}_t \in \mathbb{R}^{2}$ denote the real linear velocity, the contact boolean of all feet, the critical links' masses (like trunk, thigh, and calf), the body friction coefficient, the center of mass in body space, and the disturbance force projection in x-o-y plane, respectively. $\boldsymbol{k}_{pd} \in \mathbb{R}^{26}$ includes the proportional gain $\boldsymbol{k}_{p}  \in \mathbb{R}^{1}$, the derivative gain $\boldsymbol{k}_{d} \in \mathbb{R}^{1}$, the motor strength of each joint $\boldsymbol{\alpha} \in \mathbb{R}^{12}$ and the motor offset of each joint $\Delta \boldsymbol{q} \in \mathbb{R}^{12}$.
\subsubsection{Action Space}
The action $\boldsymbol{a}_t\in\mathbb{R}^{12}$ represents the desired increment of the joint angle w.r.t the initial pose $\boldsymbol{\mathring{q}}$, i.e. $\boldsymbol{q}^{*}_{t} = \boldsymbol{\mathring{q}} + \boldsymbol{a}_{t}$. The final desired angle $\boldsymbol{q}^{*}_{t}$ is tracked by the torque generated by the joint-level proportional–derivative (PD) controller of the actuation module in the simulator.

\subsection{Neural Network design}\label{Terrain-aware}
As shown in Fig. \ref{pics_framework}, the MAstering Risky Gap (MARG) controller consists of four sub-networks: an actor net, a critic net, an estimator net, and an elevation net. Each component is crucial in enabling the controller to adapt in real-time to the challenges posed by risky gap terrains. Next, we will discuss each part of the framework in detail. 

\subsubsection{Learn to Extract the Critical Features} The ability of our controller to master risky terrains can be attributed to two networks: an elevation net that extracts elevation features $\boldsymbol{e}^h_{t} \in \mathbb{R}^{16}$ and an estimator net that predicts the critical privilege features $\boldsymbol{e}^o_{t} \in \mathbb{R}^{7}$. These features are utilized by actor and critic networks to make informed decisions, ensuring locomotion safely and efficiently under risky terrains. 

The elevation net $\boldsymbol{E}_{\boldsymbol{\theta_1}}$ is designed to extract the elevation features $\boldsymbol{e}^h_{t}$ from the robot-centered area of $1.6$ m × $1.0$ m, which enable the robot to gain a comprehensive understanding of the height variations and contours, thereby facilitating its identification of safe zones. By incorporating the $\boldsymbol{e}^h_{t}$ into the learning process, the robot can better adapt to risky environments and enhance its locomotion ability. This network employs a Multilayer Perceptron (MLP) architecture to extract the terrain features $\boldsymbol{e}^h_{t}$. Specifically, we utilize the relative height map $\boldsymbol{\hat{h}}_t$ between the robot's base $\boldsymbol{z}_{t}$ and the surrounding terrain 
$\boldsymbol{h}_{t}$ as input, as follows:
\begin{equation}
    \boldsymbol{e}^h_{t}  = \boldsymbol{E}_{\boldsymbol{\theta_1}}(\boldsymbol{\hat{h}}_t)
 = \boldsymbol{E}_{\boldsymbol{\theta_1}}(\boldsymbol{z}_{t}-\boldsymbol{h}_{t}).
\end{equation}

The estimator net $\boldsymbol{E}_{\boldsymbol{\theta_2}}$ aims to process the history of observations $\boldsymbol{\widetilde{o}}_t^H$ and extract the critical privilege features $\boldsymbol{e}^o_{t}$ (including the estimated linear velocity $\boldsymbol{\hat{v}}_t$ and the estimated contact boolean of all feet $\boldsymbol{\hat{c}}_t$), which plays a vital role in enhancing the robot's stability and robustness during locomotion on risky terrains, particularly when encountering gaps or obstacles.
\begin{equation}
     \boldsymbol{e}^o_{t} = \boldsymbol{E}_{\boldsymbol{\theta_2}}(\boldsymbol{\widetilde{o}}^{\boldsymbol{H}}_t) 
   = (\boldsymbol{\hat{v}}_t, \boldsymbol{\hat{c}}_t)
\end{equation}
\subsubsection{Asymmetric Actor-Critic}
Certain privileged information, in addition to terrain maps, can significantly enhance the stability and robustness of the robot's locomotion ability under risky gaps. Thus, our controller adopts the asymmetric actor-critic structure for terrain-aware robot learning, enabling more effective exploration under high-dimensional spaces of legged robots. The actor and critic learn from distinct objectives: the former aims to maximize expected rewards, while the latter minimizes the difference between predicted and actual values.

The actor net is designed to derive the action $\boldsymbol{a}_t$ at each time step, which integrates several inputs to create a comprehensive representation of the current state. By leveraging multiple sources of information, the actor net can better adapt its actions to dynamic environmental conditions, allowing it to make informed decisions. The combined input vector $\boldsymbol{o}_t$ includes the proprioception observation $\boldsymbol{\widetilde{o}}_t$, the critical privilege features $\boldsymbol{e}^o_{t}$, and the elevation features $\boldsymbol{e}^h_{t}$, defined as follows:
\begin{equation}
    \boldsymbol{o}_t = \left[ \boldsymbol{\widetilde{o}}_t, \ \boldsymbol{e}^{o}_t,  \ \boldsymbol{e}^h_t \right]
\end{equation}

Except for the proprioception observation and the elevation features, the critic net also integrates more ground-truth physical knowledge $ \boldsymbol{\widetilde{s}}_t$ about the environment, which enables this net to more accurately assess the value of the current state and provide useful feedback to the actor net. The input vector $\boldsymbol{s}_t$ of the value net can be organized as follows:
\begin{equation}
     \boldsymbol{s}_t = \left[\boldsymbol{\widetilde{o}}_t, 
    \ \boldsymbol{\widetilde{s}}_t, 
    \ \boldsymbol{e}^h_t \right].
\end{equation}
\subsubsection{Concurrent Training}
These four networks in MARG are trained together by Proximal Policy Optimization ~\cite{schulman2017proximal} in simulation to achieve real-time adaptation in risky gap terrains. During training, the policy gradient $loss_{policy}$ can be backpropagated through the actor net to update its parameters of the estimator and elevation net, allowing them to learn from the rewards received during interactions with the environment. Simultaneously, the value function $loss_{value}$ is computed to minimize the error between the predicted value and the actual returns, which is also backpropagated to update the critic net. Meanwhile, the training of the estimator net is not independent of the training of the actor net, the parameters of the estimator net are also updated via a regression loss $loss_{reg}$ to reduce the Mean Squared Error (MSE), as follows:
\begin{equation}  \label{equ: loss_for_reg}
 loss_{reg} = MSE(\hat{\boldsymbol{v}}_t, \boldsymbol{v}_t) + MSE(\hat{\boldsymbol{c}}_t, \boldsymbol{c}_t)
\end{equation}

\subsection{Rewards for Footholds}
\begin{figure}[h]
\centering
\includegraphics[width=0.3\textwidth, trim=0 0 0 0,clip ]{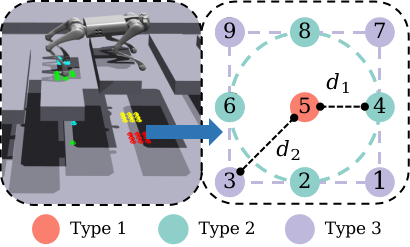}
   \caption{The feet center reward calculation: points around the robotic foot are categorized as Type 1 (at contact), Type 2 (within $d_1$ radius), and Type 3 (within $d_2$ radius).}
   \label{pics:feet_reward}
\end{figure}
The reward functions are designed to track the command velocity, penalize the unsmoothness of robot locomotion, avoid collisions, constrain the joint motion of quadrupeds, and optimize footholds, as shown in Table \ref{reward_function}. To enable safe footholds on risky terrains, we design three foot-related rewards: feet air time, feet stumble, and feet center.

Feet air time encourages the leg lifting time during movement, helping the robot cross uneven terrain. We calculate it as $\sum{_{f=0}^{4}}(\boldsymbol{t}_{air, f}-0.5)$, where $\boldsymbol{t}_{air, f}$ represents the duration each foot stays in the air (i.e., the time when it is not in contact with any terrain).

Feet stumble is a penalty for feet hitting vertical surfaces, which can prevent the robot from getting stuck or tripping during its movement. In the Isaac Gym simulator \cite{makoviychuk2021isaac}, we can easily obtain the contact force of each body at any time. By extracting the corresponding indices of each foot, we can accurately obtain the contact forces $\boldsymbol{f}_{x, y, z}=[\boldsymbol{f}_x, \boldsymbol{f}_y, \boldsymbol{f}_z]$ in the $x$, $y$, and $z$ directions. Subsequently, we use these components to define the reward, as follows: $ \text{any} \left( \| \boldsymbol{f}_{x,y} \| > 4 \cdot |\boldsymbol{f}_z| \right)$, where $\| \boldsymbol{f}_{x,y} \|$  represents the magnitude of the horizontal contact force and $|\boldsymbol{f}_z|$ denotes the absolute value of the vertical contact force.

The feet center imposes a penalty for each step on the edges, helping the robot to choose safer areas for locomotion. Specifically, we select 9 points around each foot end, as shown in Fig. \ref{pics:feet_reward}.  
We classify the points around the foot into three types based on their distance from the foot. The contact position of the foot (i.e., $id = 5$) is classified as type 1. Points within a radius  $d_1=5$ cm from the foot are type 2, while points within a circle of radius $ d_2 = \sqrt {50}$ cm from the foot end position are type 3. We can determine whether the foot is on the edge based on the heights of the points in each type, as follows: 
\begin{equation}
\begin{aligned}
n_t^{i} &= \begin{cases}
1, & \text{if } id \in  \text{Type $i$} \text{ and } h_{id} < -0.2 \\
0, & \text{else}
\end{cases} 
\label{equ: foot_reward}
\end{aligned}
\end{equation}
where $id$ and $h_{id}$ denote the index and height of this point.
\begin{table}[h]
\caption{Reward Terms.}
\label{reward_function}
\setlength\tabcolsep{2pt} 
\begin{center}
\begin{tabular}{c|c|c|c}
\hline
\textbf{Term} & \textbf{Reward} & \textbf{Equation}  & \textbf{Weight} \\
\hline
\multirow{2}*{Task} & Lin. velocity tracking &  $e^{-4||\boldsymbol{v}_{xy}^{*}-\boldsymbol{v}_{xy}||^2}$   & 1.0\\
     & Ang. velocity tracking &  $e^{-4(\boldsymbol{w}_{yaw}^{*}-\boldsymbol{w}_{yaw})^2}$    & 0.5\\
\hline
\multirow{5}*{Smoothness} 
    & Linear velocity ($z$) & $\boldsymbol{v}_z^2$   & -2.0\\
    & Angular velocity ($xy$) & $||\boldsymbol{w}_{xy}||^2$ & -0.05\\
    & Joint torque & $||\boldsymbol{\tau}||^2$  &  $-e^{-5}$\\
    & Action rate & $||\boldsymbol{a}_{t}-\boldsymbol{a}_{t-1}||^2$ & -0.01\\
    & Joint accelerations & $||\ddot{\boldsymbol{q}}||^2$ & $-2.5  e^{-7}$\\
\hline
Safety &Collisions& $-n_{collision}$ & 1.0\\
\hline
\multirow{2}*{Pose}  &Orientation & $||\boldsymbol{q}_{xy}||^2$ & -0.2\\
 & Joint motion limit& $\sum{_{j=0}^{12}}||\boldsymbol{q}_{t, j}-\mathring{\boldsymbol{q}}_{j}||$ & -0.02\\
\hline
\multirow{3}*{Footholds}
& Feet air time& $\sum{_{f=0}^{4}}(\boldsymbol{t}_{air, f}-0.5)$ & 1.0\\
& Feet stumble& $ \text{any} \left( \| \boldsymbol{f}_{x,y} \| > 4 \cdot |\boldsymbol{f}_z| \right)$ & -1.0\\
& Feet center& $ c_t \cdot (n_t^2 + 2*n_t^3)$ & -0.01 \\
\hline
\end{tabular}
\end{center}
\end{table}


\section{Terrain Map Generation Model}
An accurate terrain map $\boldsymbol{\hat{h}}_t$ is critical for the successful deployment of our controller in real-world environments. Previous approaches have relied heavily on accurate global pose estimation \cite{wang2024sftimsimpleframeworkenhancing} and clean point cloud inputs \cite{zhang2023learning} to accumulate elevation information. Furthermore, current deployment solutions typically require multiple external sensors, such as arrays of cameras or LiDAR systems, which substantially increase the complexity and cost of practical implementation \cite{jenelten2024dtc} \cite{fankhauser2016universal}. 

In this section, we only use one LiDAR sensor for the perception of risky terrains, which significantly reduces the complexity of hardware deployment. However, there are several challenges with LiDAR-based localization and mapping. First, typical LiDAR-based localization methods often suffer from height drift, which can be detrimental for quadrupedal locomotion as the height estimate is used to construct a height map. Furthermore, LiDAR-based mapping maintains a local map by employing a global sliding window, which is computationally intensive. 

To address these limitations, we propose a Terrain Map Generation (TMG) model that incorporates kinematic measurements and a hash-based local map to provide accurate and computationally efficient elevation information. In the following sections, we will elaborate on the key components of our model.


\subsection{Contact-aware State Estimation}
We first estimate the transformation between two neighboring frames using an Error State Kalman Filter (ESKF) approach, which will be used later for constructing an ego-centric elevation map. Existing LiDAR-based localization methods \cite{xu2022fast,shan2020lio} have drift issues and cannot be directly implemented. To address these, we incorporate the contact point positions $\boldsymbol{p}_{f_i} \in \mathbb{R}^3$ in the state vector. 

The ground-truth state is defined as a tuple $\boldsymbol{x} \in SO(3) \times \mathbb{R}^{27}$ comprising the following states:
\begin{equation}
\boldsymbol{x} = 
\begin{bmatrix}
\boldsymbol{R}_{wb} & \boldsymbol{p}_{wb}&  \boldsymbol{v}_{wb} & \boldsymbol{b}_a &  \boldsymbol{b}_{\omega}&  \boldsymbol{p}_{f_1}...\boldsymbol{p}_{f_4} & \boldsymbol{g} 
\end{bmatrix}
\end{equation}
where $\boldsymbol{R}_{wb}\in SO(3)$ denotes the rotation matrix from the body frame to the world frame, while $\boldsymbol{p}_{wb} \in \mathbb{R}^3$ and $\boldsymbol{v}_{wb} \in \mathbb{R}^3$ represent the position and velocity of the body frame in the world frame, respectively. The terms $\boldsymbol{b}_a$ and $\boldsymbol{b}_{\omega}$ correspond to accelerometer and gyroscope biases, while $\boldsymbol{g}$ represents the gravity vector in the world frame that requires initialization. To incorporate the kinematic properties of the robot into the estimation, we augment the state vector with the positions of contact points $\boldsymbol{p}_{f_i}$, enabling us to mitigate positional drift in the estimation process.
\subsubsection{Forward Propagation} 
Considering the state transition from time step $t$ to $t+1$ and neglecting the noise terms ( $\boldsymbol{n}_a$, $\boldsymbol{n}_{\omega}$, and $\boldsymbol{n}_{p_{fi}}$), the forward propagation can be governed by:
\begin{equation}
    \boldsymbol{\bar{x}}_{t+1} = \boldsymbol{\bar{x}}_t \boxplus (\Phi(\boldsymbol{\bar{x}}_t, \boldsymbol{u}_t, 0) \Delta t)
\end{equation}
\begin{equation}
    \Phi(\boldsymbol{\bar{x}}_t, \boldsymbol{u}_t, \boldsymbol{n}_t) = \begin{bmatrix}
\boldsymbol{\omega}_{m_t} -\boldsymbol{b}_{\omega_t} - \boldsymbol{n}_{\omega_t} \\
\boldsymbol{v}_{wb_t} \\
\boldsymbol{R}_{wb_t} (\boldsymbol{a}_{m_t} - \boldsymbol{b}_{a_t} - \boldsymbol{n}_{a_t}) + \boldsymbol{g}_t \\
\boldsymbol{n}_{ba_t} \\
\boldsymbol{n}_{b\omega_t} \\
\boldsymbol{n}_{p_{f_{i, t}}} \\
0_{3 \times 1}
\end{bmatrix}
\end{equation}
where $ \boldsymbol{\bar{x}_t}$ represents the propagated state and $\Phi$ denotes the forward propagation function. The IMU measurements $\boldsymbol{u}$ are influenced by Gaussian noise terms: $\boldsymbol{n}_{a}$ for acceleration and $\boldsymbol{n}_{\omega}$ for angular velocity. During the swing phase, the foot position is affected by white noise $\boldsymbol{n}_{p_{f_i}}$. Due to the lack of ground contact, the uncertainty in foot position increases significantly. This is typically modeled by assigning a large variance to this process noise.



Let the measurement input vector be $\boldsymbol{u}=\left[ \boldsymbol{\omega}_{m} \ \boldsymbol{a}_{m} \right]$ and the noise vector be $\boldsymbol{n} = \left[\boldsymbol{n}_{a} \ \boldsymbol{n}_{\omega} \ \boldsymbol{n}_{b\omega} \ \boldsymbol{n}_{ba} \ \boldsymbol{n}_{p_{f_i}} \right]$. The components $\boldsymbol{n}_{b\omega}$ and $\boldsymbol{n}_{ba}$ correspond to the random walk noise for the IMU biases $\boldsymbol{n_{\omega}}$ and $\boldsymbol{n_{a}}$, respectively. Furthermore, the error state $\boldsymbol{\widetilde{x}}$ is characterized by:
\begin{equation}
\boldsymbol{\widetilde{x}} = \boldsymbol{x} \boxminus \boldsymbol{\bar{x}} \approx \boldsymbol{F}_{\widetilde{x}} \boldsymbol{\widetilde{x}} + \boldsymbol{F}_{n} \boldsymbol{n}
\end{equation}
where $\boldsymbol{F}_{\widetilde{x}}$ and $\boldsymbol{F}_{n}$ represent the Jacobian matrices of $\Phi$ with respect to $\widetilde{\boldsymbol{x}}$ and ${\boldsymbol{n}}$, respectively.

\subsubsection{LiDAR Measurement} For every LiDAR points $^{L}\boldsymbol{p}_{j}$, we model the LiDAR measurement $\boldsymbol{h}_j(\boldsymbol{x}_t, ^{L}\boldsymbol{p}_{j}+^{L}\boldsymbol{n}_j)$ as:
\begin{equation}
    0 = \boldsymbol{u}^T_j (^G\boldsymbol{T}_{I}\ ^{I}\boldsymbol{T}_{L}(^{L}\boldsymbol{p}_{j} + ^{L}\boldsymbol{n}_{j}) - ^{G}\boldsymbol{q}_j)
\end{equation}
where $\boldsymbol{u}_j$ denotes the unit vector of the corresponding plane, $^G\boldsymbol{T}_{I}$ represents the transformation from the IMU frame to the world frame, where the rotation component $^G\boldsymbol{T}_I$ equals $\boldsymbol{R}_{wb}$ since we assume that the IMU frame coincides with the body frame, and $^{I}\boldsymbol{T}_{L} = (^{I}\boldsymbol{R}_{L}, ^I\boldsymbol{t}_L)$ is the extrinsic transformation from LiDAR frame to IMU frame. $^G\boldsymbol{q}_j$ represents a point on the corresponding plane, and $^{L}\boldsymbol{n}_j$ denotes the LiDAR measurement noise. The linearized measurement model is given by:
\begin{equation}
0 \simeq \boldsymbol{h}_j(\bar{\boldsymbol{x}}_t, \boldsymbol{0}) + \boldsymbol{H}_{j} \widetilde{\boldsymbol{x}}_t + \boldsymbol{n}_j
\end{equation}
\begin{equation}
\begin{aligned}
\boldsymbol{H}_{j} &= \left.\frac{\partial \boldsymbol{h}_j(\bar{\boldsymbol{x}}_t \boxplus \widetilde{\boldsymbol{x}}_t, \boldsymbol{0})}{\partial \widetilde{\boldsymbol{x}}_t}\right|{\widetilde{\boldsymbol{x}}_t = \boldsymbol{0}} \\
&= \boldsymbol{u}^T_j \begin{bmatrix} \begin{smallmatrix}
-{\boldsymbol{\bar{R}}}_{wb} ( ^{I}\boldsymbol{R}_{L} {^{L}}\boldsymbol{p}_{j} + ^I\boldsymbol{t}_L ) ^{\wedge} & \boldsymbol{I}_{3 \times 3} & \boldsymbol{0}_{3 \times 21}\end{smallmatrix}
\end{bmatrix}
\end{aligned}
\end{equation}
\noindent where $\boldsymbol{H}_j$ is the Jacobian matrix of $\boldsymbol{h}_j$ with respect to $\widetilde{\boldsymbol{x}}$, and $\boldsymbol{n}_j \sim \mathcal{N}(\boldsymbol{0}, \boldsymbol{R}_j)$ represents the LiDAR measurement noise.
\subsubsection{Kinematics Measurement} The kinematics measurement can be defined as:
\begin{equation}
\boldsymbol{p}_{{f_{i}^{rel}}} = \boldsymbol{R}_{wb}^T \cdot (\boldsymbol{p}_{wb} - \boldsymbol{p}_{f_i})
\end{equation}
%
where the term $\boldsymbol{p}_{{f_{i}^{rel}}}$ represents the measured relative position of the foot contact point $\boldsymbol{p}_{f_i}$ in the body frame.
Assuming that there is no relative sliding between the contact point and the ground, the velocity of the foot can be assumed as zero, and the residuals of contacted foot velocity $\boldsymbol{h}_{cv}$ and position  $\boldsymbol{h}_{cp}$ can be defined as:
\begin{equation}
\boldsymbol{h}_{cv} = \boldsymbol{v}_{wb} + \boldsymbol{R}_{wb} \cdot (\boldsymbol{v}_{f_{i}^{rel}} + (\boldsymbol{\omega}_{m}-\boldsymbol{b}_{\omega}) \times \boldsymbol{p}_{f_i^{rel}})
\end{equation}
\begin{equation}
\boldsymbol{h}_{cp} = \boldsymbol{p}_{f_{i}^{rel}} - \boldsymbol{R}_{wb}^T \cdot (\boldsymbol{p}_{wb}-\boldsymbol{p}_{f_i})
\end{equation}
\begin{equation}
\begin{aligned}
\boldsymbol{H}_{cv} &= \left.\frac{\partial \boldsymbol{h}_{cv}(\bar{\boldsymbol{x}}_t \boxplus \widetilde{\boldsymbol{x}}_t, \boldsymbol{0})}{\partial \widetilde{\boldsymbol{x}}_t}\right|{\widetilde{\boldsymbol{x}}_t = \boldsymbol{0}} \\
&= \left[\begin{array}{cc}
 - \boldsymbol{\bar{R}}_{wb} \cdot (\boldsymbol{v}_{f_i^{rel}} + (\boldsymbol{\omega}_{m}-\boldsymbol{\bar{b}}_{\omega})^{\wedge} \boldsymbol{p}_{f_i^{rel}})^{\wedge} & 
 \boldsymbol{0}_{3 \times 3} 
\end{array} \right. \\
 &\qquad \left. \begin{array}{cccc}
 \boldsymbol{I}_{3 \times 3} & \boldsymbol{0}_{3 \times 3} &\ \boldsymbol{\bar{R}}_{wb} \cdot \boldsymbol{p}_{f_i^{rel}}^{\wedge} & \boldsymbol{0}_{3 \times 12} 
\end{array} \right]
\end{aligned}
\end{equation}
\begin{equation}
\begin{aligned}
\boldsymbol{H}_{cp} &= \left.\frac{\partial \boldsymbol{h}_{cp}(\bar{\boldsymbol{x}}_t \boxplus \widetilde{\boldsymbol{x}}_t, \boldsymbol{0})}{\partial \widetilde{\boldsymbol{x}}_t}\right|{\widetilde{\boldsymbol{x}}_t = \boldsymbol{0}} \\
& = \begin{bmatrix} \boldsymbol{\bar{R}}_{wb} \cdot (\boldsymbol{p}_{f_i^{rel}})^{\wedge}& -\boldsymbol{I}_{3 \times 3} & \boldsymbol{0}_{3 \times 9}& \boldsymbol{I}_{3 \times 12}
\end{bmatrix}
\end{aligned}
\end{equation} 
where $\boldsymbol{H}_{cv}$ and $\boldsymbol{H}_{cp}$ are the Jacobian matrices. $\boldsymbol{v}_{f_{i}^{rel}}$ denotes the measured relative foot velocity. The terms $\boldsymbol{\bar{R}}_{wb}$ and $\boldsymbol{\bar{b}}_{w}$ represent the propagated state of rotation from the body frame to the world frame and IMU gyroscope bias.

\subsubsection{State Update} 
Following the conventional IESKF process \cite{xu2022fast}, we formulate the state estimation problem as a Maximum A Posteriori (MAP) estimation:
\begin{equation}
\begin{aligned}
\text{minimize}_{\widetilde{\boldsymbol{x}}_t} ( & \| \boldsymbol{x}_t \boxminus \boldsymbol{\bar{x}}_t \|^2_{\boldsymbol{\bar{P}}^{-1}_t} + \sum_{j=1}^{m} \| \boldsymbol{h}_j + \boldsymbol{H}_j \boldsymbol{\tilde{x}}_t \|^2_{\boldsymbol{R}^{-1}_j} \\
& + \| \boldsymbol{h}_{cv} (\boldsymbol{\bar{x}}_t, \boldsymbol{u}_t, 0, 0) + \boldsymbol{H}_{cv} \boldsymbol{\tilde{x}}_t \|^2_{\boldsymbol{\Sigma}_{cv}^{-1}} \\
& + \| \boldsymbol{h}_{cp} (\boldsymbol{\bar{x}}_t, \boldsymbol{u}_t, 0) + \boldsymbol{H}_{cp} \boldsymbol{\tilde{x}}_t \|^2_{\boldsymbol{\Sigma}_{cp}^{-1}} )
\end{aligned}
\end{equation}

The optimal state estimate is obtained through the Kalman gain $\boldsymbol{K}_t$ and the corresponding state and covariance updates \cite{xu2022fast}. Finally, the neighboring state 
$\mathring{\boldsymbol{x}}_{t}$ is updated through:
\begin{equation}
\mathring{\boldsymbol{x}}_{t} =  \Phi(\boldsymbol{\bar{x}}_t, \boldsymbol{u}_t, 0) \Delta t \boxplus \boldsymbol{K}_{t} \begin{bmatrix} \boldsymbol{h}_1 \ \hdots \ \boldsymbol{h}_m \ \boldsymbol{h}_{cv} \ \boldsymbol{h}_{cp} \end{bmatrix}^T
\end{equation}
This updated state estimate can subsequently be utilized to refine the local map representation.

\subsection{Ego-centric Elevation Mapping}
Conventional elevation mapping maintains a global sliding window tracking absolute poses relative to a fixed origin, updating the entire map as the robot moves. This global approach has two key limitations: high computational overhead and deteriorating reliability with global odometry errors \cite{Ren_2024}.

Confronting these challenges, we propose an ego-centric mapping strategy focusing on the robot's immediate vicinity. Using the local state estimate $\boldsymbol{\mathring{x}}_t$, our approach enables precise local updates while reducing both computational cost and dependency on global pose estimation accuracy.

\subsubsection{Local Map Sliding}
We employ a hashmap-based approach for zero-copy local map sliding, maintaining a local elevation map of dimensions $\boldsymbol{L} \in \mathbb{R}^3$ with resolution $\boldsymbol{r}$, discretized into $\boldsymbol{N}$ cells. Each cell is referenced by its global index $\boldsymbol{g}_i = (g_{i_x}, g_{i_y}, g_{i_z})$, which is hashed based on its position relative to the body frame. We also maintain a local index $\boldsymbol{l}_i = (l_{i_x}, l_{i_y}, l_{i_z})$ within the map dimensions.

As the robot traverses the environment, we update the global indices according to:
\begin{equation}
\boldsymbol{g}_i = \boldsymbol{\mathring{R}}_{wb} \boldsymbol{N}(\boldsymbol{g}_i) + \boldsymbol{\mathring{p}}_{wb} \oslash \boldsymbol{r}
\end{equation}
where $\mathring{\boldsymbol{R}}_{wb} \in SO(3)$ and $\mathring{\boldsymbol{p}}_{wb}\in\mathbb{R}^3$ denote the incremental rotation and translation from the body frame to the world frame, and $\oslash$ represents the element‑wise division operator. The local indices are obtained by normalizing the global indices to ensure they remain within the elevation map:
\begin{equation}
\boldsymbol{l}_i = \text{normalize}(\boldsymbol{g}_i, \boldsymbol{L})
\end{equation}
At each time step, the grid cells are incrementally updated using LiDAR measurements to maintain an accurate representation of the local environment.

\subsubsection{Local Map Updating}
We implement a systematic approach for updating the local map grid to handle the beam divergence characteristics of LiDAR sensors. Initially, we employ Statistical Outlier Removal (SOR) to filter measurement noise from the point cloud data.

Then, we maintain a cached frame $\boldsymbol{C}$ to track grid occupancy states through ray casting \cite{han2019fiesta} and store probabilistic occupancy information. For a point $\boldsymbol{p}$ in the LiDAR frame where $\boldsymbol{x}_p \in \mathbb{R}^3$, and its corresponding grid cell $\boldsymbol{g}_p$ in the local map, we update the occupancy probability using a logarithmic odds formulation:
\begin{equation}
\begin{split}
    \boldsymbol{C}{pro|_t} = \boldsymbol{C}{pro|_{t-1}} + n_{hit} \log\left(\frac{p_{hit}}{1 - p_{hit}}\right) +  \\ n_{miss} \log\left(\frac{p_{miss}}{1 - p_{miss}}\right)
\end{split}
\end{equation}
where $n_{hit}$ and $n_{miss}$ represent the number of hit and miss points within the grid cell, respectively. $p_{hit}$ and $p_{miss}$ denote their corresponding probability parameters.

To effectively manage both dynamic and static objects in the environment, we implement probability bounds using lower $T_{low}$  and upper $T_{high}$ thresholds:
\begin{equation}
\boldsymbol{C}{pro|_t} = \max(\min(\boldsymbol{C}{pro|_t}, T_{high}), T_{low})
\end{equation}

This bounded update ensures robust occupancy estimation while maintaining adaptability to environmental changes.

\subsubsection{Priori Ray Interpolation}

To extract a relative terrain map $\boldsymbol{\hat{h}}_t$, we determine the highest occupied voxel height, denoted as $\boldsymbol{\hat{h}}(x_i, y_i)$, within each column of the occupancy grid representation's horizontal plane. Moreover, we implement a bidirectional ray-based interpolation strategy to replenish regions with missing data, where empty columns are assigned elevation values through both forward and reverse traversals. Specifically, during forward traversal, the elevation value $\boldsymbol{\hat{h}}(x_f, y_f)$ is obtained from the farthest detected occupied cell at distance $d_{far}$, while in reverse traversal, the elevation value $\boldsymbol{\hat{h}}(x_n, y_n)$ is taken from the nearest occupied cell at distance $d_{near}$. As depicted in Fig.~\ref{pics_framework}, this dual-direction interpolation effectively maintains terrain continuity while preserving critical structural features. Finally, the relative terrain map $\boldsymbol{\hat{h}}_t$ is fed into the elevation net to generate the elevation feature $\boldsymbol{e}_t^h$.
\begin{figure}[H]
   \centering
   \includegraphics[width=0.46\textwidth, trim=1 10 1 10,clip]{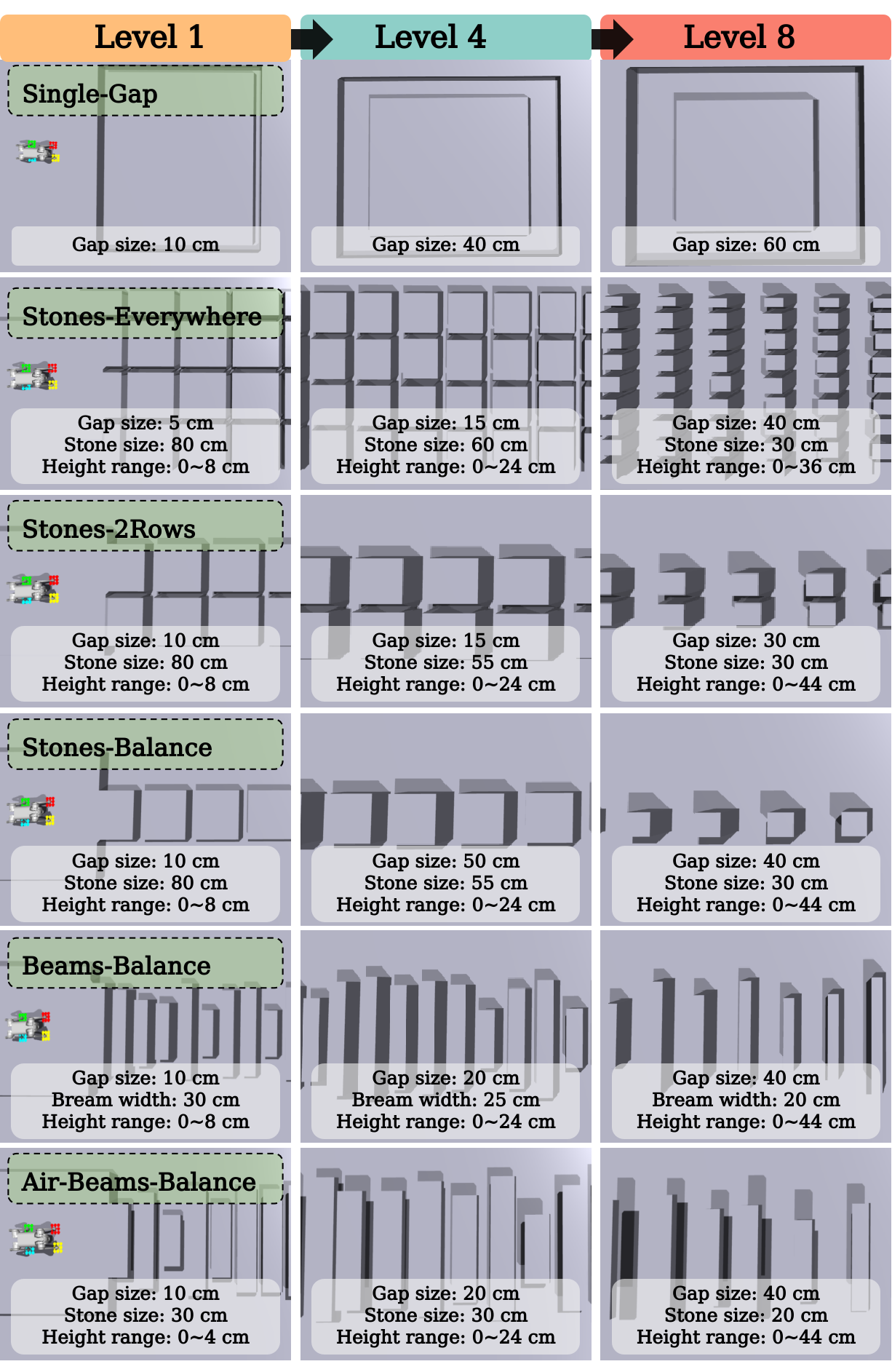}
   \caption{The gap terrains of 8 different levels used in the experiments, such as Single-Gap, Stones-Everywhere, Stones-2Rows, Stones-Balance, Breams-Balance, and Air-Beams-Balance terrains. As the level increases, the gaps are enlarged, and the width and height of stones or beams are adjusted.}
   \label{pics:risky_terrain}
\end{figure}

\section{Experimental Steup}\label{Experiments}
\subsection{Simulation}
In this paper, we trained 4096 environments using Unitree Go1 with diverse domain randomization in parallel on the Isaac Gym simulator \cite{makoviychuk2021isaac} for 20000 episodes, and each episode was terminated and reset under specific criteria, which included collisions of the robot's trunk or hips with the terrain, or the height of each foot falling below -0.2 m, or being trapped for 20 seconds. The randomized parameters of the environment are randomized at the initialization stage, as shown in Table \ref{tab:dom}. We utilized a game-inspired curriculum \cite{wang2021survey} to ensure progressive locomotion policy learning over risky terrains, as shown in Fig. \ref{pics:risky_terrain}. The terrains consisted of Single-Gap, Stones-Everywhere, Stones-2Rows, Stones-Balance, Breams-Balance, and Air-Beams-Balance terrains with 8 different levels. As the level increased, we progressively enlarged the gaps while reducing the width and elevating the height of the stones or beams. 

\begin{table}[t]
\caption{The Randomization range of parameters.}
\label{tab:dom}
\begin{center}
\begin{tabular}{c c c}
\hline
\textbf{Parameters} & \textbf{Range}  & \textbf{Unit} \\
\hline
  $\boldsymbol{K}_{p}$ factor &   [0.9, 1.1] & Nms/rad\\
  $\boldsymbol{K}_{d}$ factor  &  [0.9, 11] & Nms/rad\\
  Payload &  [-1.0, 2.0) & kg\\
  System delay   &  [0.0, 12] & ms\\
  Frictions coefficient  &  [0.2, 1.25) & - \\
  Center of mass shift &  [-0.05, 0.05) & m\\
  Motor strength factor &  [0.9, 1.1] & Nm \\
   Noise of the terrain map $h_t$ &  [-5, 5) & cm \\
   \hline
\end{tabular}
\end{center}
\end{table}

The actor and critic networks for all controllers were trained together using PPO ~\cite{schulman2017proximal}, and the architectures and the key parameter settings (such as clipping range, learning rate, etc.) were listed in ~\cite{rudin2022learning}. In contrast, the architecture of the Estimator and Elevation networks of MARG were $[258, 128, 7]$ and $[128, 64, 16]$, respectively, utilizing Rectified Linear Units (ReLU) as the activation function. The entire training was performed on a desktop PC with Intel (R) Xeon(R) Platinum 8370C CPU @ 2.80GHz, 80 GB RAM, and an NVIDIA A100 GPU. Training of the MARG algorithm cost approximately twelve hours.

\subsection{Hardware} 
To validate our approach, we conducted extensive real-world trials using Unitree Go1 and Go2 quadruped robots. Both platforms were equipped with a Livox Mid360 LiDAR operating at 10 Hz, while maintaining a consistent 50 Hz synchronization frequency across the control policy, state estimator, and elevation mapping network. All joint commands were transmitted through an Ethernet interface for reliable communication. The Go1 utilized an onboard Intel NUC for state estimation and control computations, adding nearly 2 kg of payload. We configured the controller with proportional and derivative gains of $\boldsymbol{k}_{p} = 30.0$ and $\boldsymbol{k}_{d} = 0.8$, respectively. In contrast, the Go2's computational tasks were handled by an externally Ethernet-connected laptop featuring an Intel i7-12700H CPU, and we used slightly higher control gains with $\boldsymbol{k}_{p} = 40.0$ and $\boldsymbol{k}_{d} = 1.2$.

\begin{table*}[h] \small
\caption{Comparison with different quadrupedal parkour locomotion controllers.}
\label{pre_controllers}
\begin{center}
\begin{tabular}{l|c|c|c|c|c}
\hline
\textbf{Controllers} & \textbf{MARG} &  \textbf{Anymal Parkour} & \textbf{PIE} & \textbf{Parkour} & \textbf{Extreme Parkour} \\
\hline
\textbf{Agent numbers} & 4096 & 4096 & 4096 & 256 & 192 \\
\hline
\textbf{GPU memory} & $\approx$  12 GB & $>$ 45 GB & $>$20 GB & $>$ 15 GB & $>$ 12 GB \\
\hline 
\textbf{Training on single phase} & Yes & No & Yes & No & No \\
\hline
\textbf{Deploy with single policy} & Yes & No & Yes & Yes & Yes \\
\hline
\textbf{Extra sensors during simulation} & $\times$ & \checkmark
 & \checkmark &\checkmark & \checkmark \\
 \hline
\textbf{Extra sensors during deployment} &1 LiDAR & 6 Depth Cameras and 1 LiDAR & \multicolumn{3}{c}{1 Depth Camera} \\
\hline
\end{tabular}
\end{center}
\end{table*}
\section{Results and Discussion}\label{Experiments}
\subsection{Algorithm Comparison}
\begin{figure}[H]
   \centering
\includegraphics[width=0.49\textwidth, trim=1 5 1 15,clip]{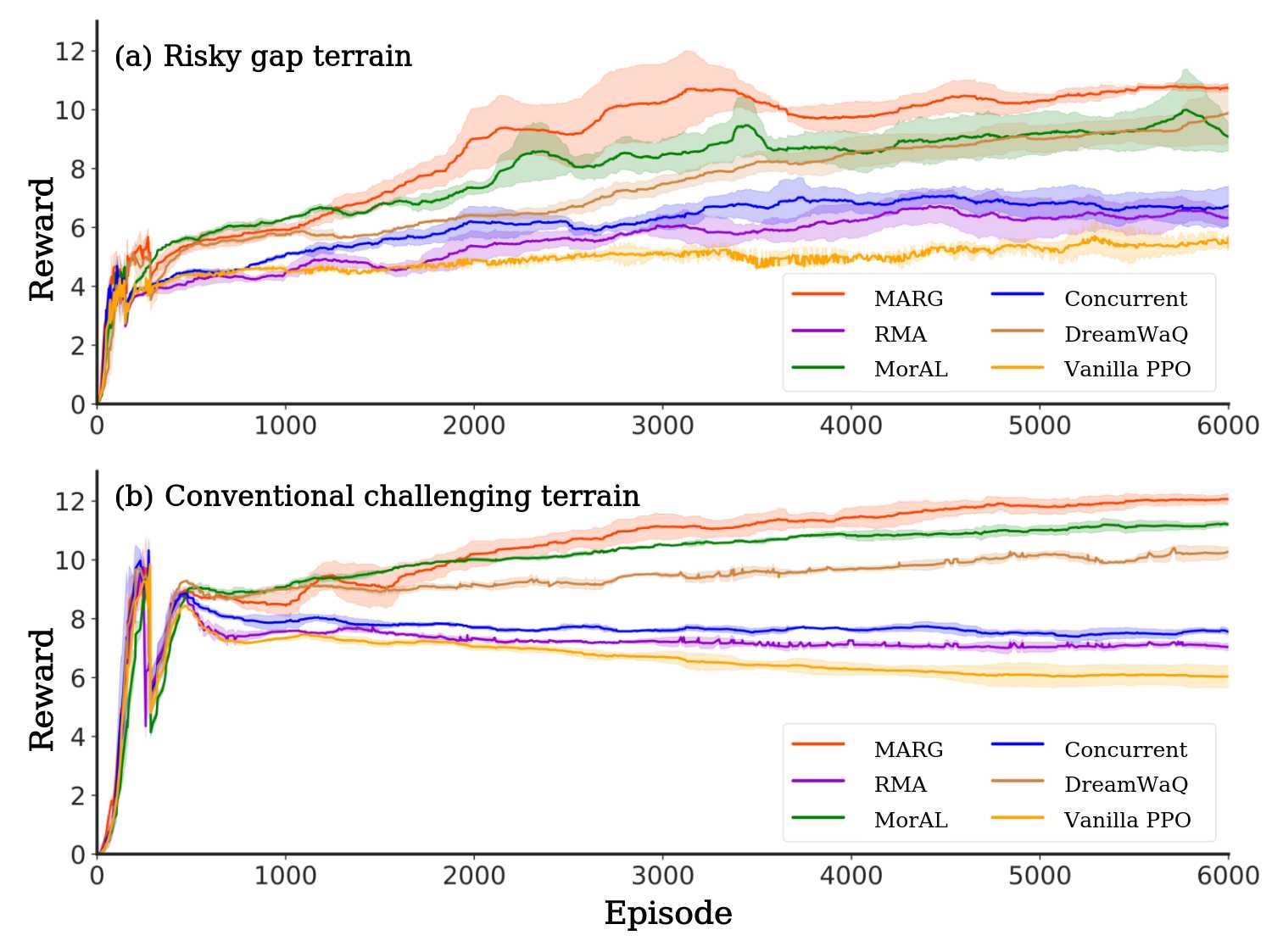}
   \caption{The average rewards of six controllers on 4096 Go1 over 6000 episodes for two different types of terrain: (a) Risky gap terrain and (b) Conventional challenging terrain. Each curve's shaded region represents the standard deviation of reward values across three different random seeds, indicating the uncertainty in the results.}
   \label{pics:compare_drl}
\end{figure}

Fig. \ref{pics:compare_drl} (a) illustrates the learning performance of six different DRL algorithms (including MARG, MorAL \cite{luo2024moral}, Vanilla PPO \cite{smith2022walk}, Concurrent \cite{ji2022concurrent}, RMA \cite{kumar2021rma}, and DreamWaQ \cite{nahrendra2023dreamwaq}) in the risky gap terrain. MARG demonstrates the most robust overall performance due to its capacity for environmental perception, resulting in reward values that significantly exceed those of the other algorithms. MorAL and DreamWaQ exhibit notable performance in the later stages, attributed to their effective exploitation mechanisms, which incorporate terrain maps into the critic network and indirectly guide policy updates. However, these algorithms are prone to converging to local optima without active terrain perception.
Due to the inclusion of some privileged information, such as mass or foot contact, the reward values of Concurrent and RMA are moderate but fluctuate greatly and clearly cannot adapt to the gap terrain. In contrast, Vanilla PPO consistently exhibits the lowest reward values, demonstrating slow growth and poor convergence. MARG shows minimal fluctuations in the later stages, indicating the best stability on gap terrain.

To further validate the generalizability of MARG, we extend the comparison to a conventional challenging terrain, consistent with the terrain settings employed in MorAL \cite{luo2024moral} and DreamWaQ \cite{nahrendra2023dreamwaq}. As shown in Figure \ref{pics:compare_drl} (b), the performance trend of the six controllers is highly consistent with that in the risky gap terrain. Specifically, MARG still exhibits the best performance, with MorAL and DreamWaQ delivering better average rewards than Concurrent, Vanilla PPO, and RMA. Overall, these experiments demonstrate that MARG has significant advantages, whether for risky or conventional challenging terrain. This is attributed to MARG not only acquiring the explicit estimation but also the terrain map $\boldsymbol{h}_{t}$ surrounding the robot, which enables the policy to reason about the robot's states. 

While several existing learning-based parkour controllers (including Anymal Parkour \cite{hoeller2024anymal}, PIE \cite{luo2024pie}, Extreme Parkour \cite{cheng2024extreme}, and Parkour \cite{zhuang2023robot}) have demonstrated capabilities in gap terrains, our comprehensive evaluation reveals MARG's distinct advantages across multiple performance dimensions, as shown in Table \ref{pre_controllers}. The TMG model enables MARG to achieve remarkable training efficiency by eliminating the dependency on exteroceptive sensors (e.g., LiDARs or cameras) during simulation training. It reduces GPU memory consumption, allowing efficient large-scale training with 4096 parallel agents with minimal resource requirements. Moreover, the asymmetric actor-critic framework enables both MARG and PIE to achieve single-stage end-to-end training, which facilitates direct sim-to-real transfer without intermediate adaptation phases. This reduces the complexity of the training process and simplifies parameter adjustment and model optimization. Furthermore, unlike Anymal Parkour \cite{hoeller2024anymal}, MARG and others only rely on a single LiDAR or depth camera for environmental perception during deployment. This simplified approach not only significantly reduces system complexity but also remarkably cuts down on hardware costs. Overall, MARG demonstrates superior training efficiency and more practical deployment features among these controllers.
 
\subsection{Explicit Estimation Comparison}
The accurate foot contact state and linear velocity are crucial for the locomotion of legged robots, yet the IMU and joint encoders of these robots cannot directly acquire such key data \cite{yoon2023invariant}. Model-based estimators \cite{bloesch2013state}, \cite{bloesch2013state1}, \cite{hartley2020contact} often rely on precise dynamics models for estimation and also fail to handle foot slip. While model-free methods \cite{lin2022legged}, \cite{buchanan2022learning}, \cite{liu2020tlio} allow robots to search for optimal strategies through continuous trial-and-error learning from data, thus avoiding the challenges of precise modeling. For example, NMN \cite{youm2024leggedrobotstateestimation} uses supervised learning to train a neural network to estimate contact probability and body linear velocity. However, this estimator is highly dependent on pre-collected labeled data. In complex and unpredictable environments such as risky gaps, data collection is extremely difficult. This leads to poor generalization performance of the model, making it hard to adapt to new or unexpected terrain features. To address these issues, algorithms like MorAL \cite{luo2024moral} and DreamWaQ \cite{nahrendra2023dreamwaq} train estimators and policies simultaneously on multiple terrains. Based on this, our MARG integrates terrain map information and privileged information during training, enabling more accurate estimation of foot contact states and linear velocities.

\begin{figure}[h]
   \centering
    \includegraphics[width=0.49 \textwidth, trim=5 5 5 10,clip]{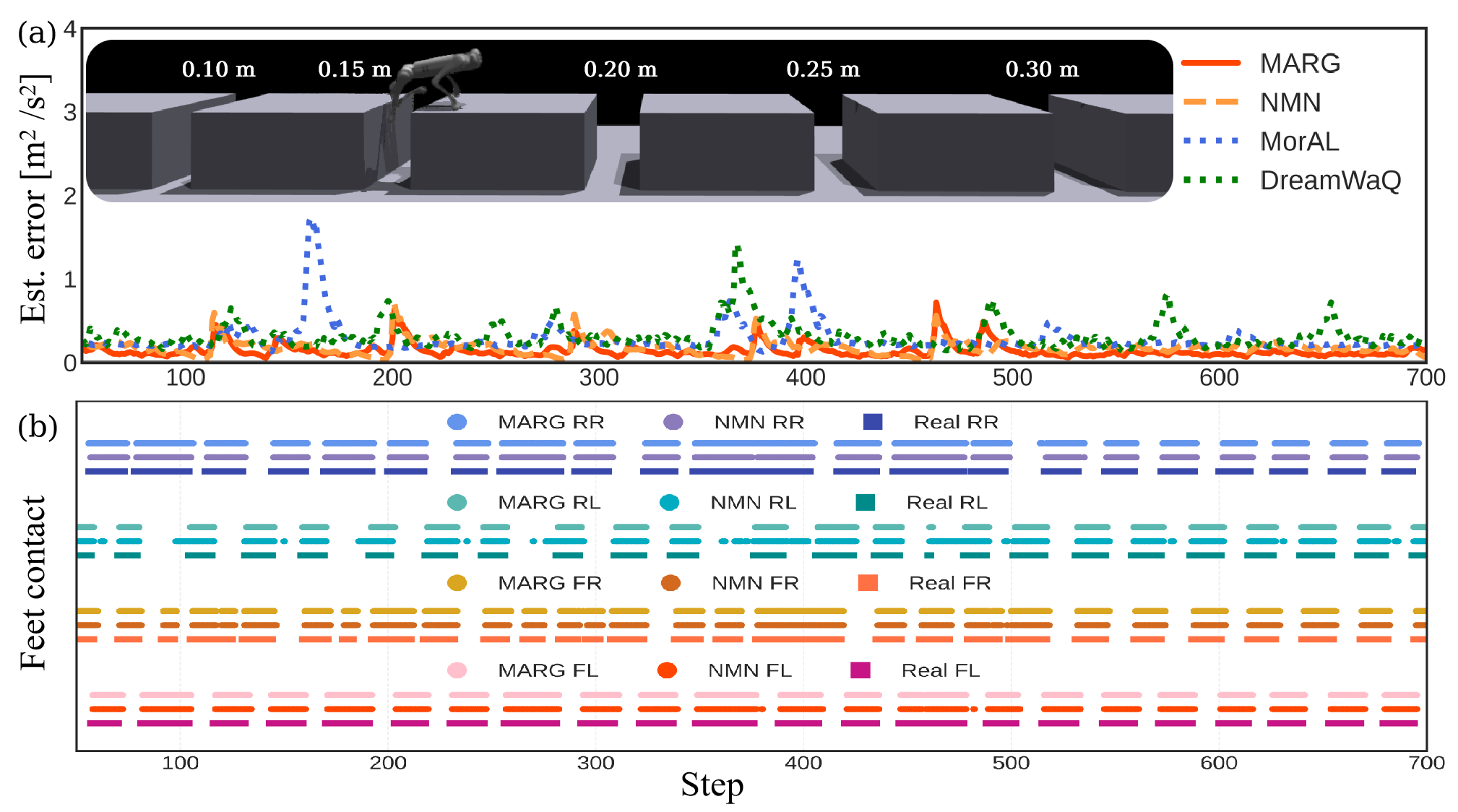}
   \caption{Accuracy of Estimator Net on various gaps ranging from 0.10 to 0.30 m. (a) The squared velocity estimation error of MARG, NMN, MorAL, and DreamWaQ. (b) Comparison of real and estimated foot contact states.}
   \label{pics:est}
\end{figure}

We further analyze the accuracy of the Estimator net on estimated linear velocity and foot contact states during Go1 across the continuous gap terrains, as shown in Fig. \ref{pics:est}. The top panel illustrates the squared estimation error across a sequence of steps for three algorithms: MARG, NMN, MorAL, and DreamWaQ. MARG consistently maintains lower estimation errors across varying gap sizes, highlighting its superior environmental perception capabilities. NMN exhibits an estimation error pattern relatively close to MARG, indicating that there are relatively minor discrepancies between them in terms of velocity estimation error. In contrast, MorAL and DreamWaQ exhibit higher peaks in estimation error, particularly during specific gap transitions, indicating challenges in adapting to abrupt environmental changes.

The bottom panel focuses on comparing real and estimated contact states in teams of the front left (FL) and front right (FR) feet, as well as the rear left (RL) and rear right (RR) feet. The estimated contact states of MARG closely match the real states, accurately predicting foot contact across all feet. Meanwhile, each foot displays similar contact patterns, with the contact states maintaining a consistent pattern and slight periodic differences between the front and rear feet. FL and RR have similar contact durations longer than FR and RL, suggesting that the robot uses a trot gait to traverse the gaps. NMN also showcases a relatively good performance in predicting contact states. Although the RL foot shows a slightly noticeable prediction error, this minor deviation does not overshadow NMN's overall capacity to approximate the real contact states.

Overall, MARG proves to be a highly effective algorithm in both linear velocity estimation and contact state prediction. Its superior environmental perception and accurate state estimation contribute to stable and efficient locomotion in risky gap terrains. While NMN shows potential with a performance close to MARG in some aspects, it relies on a learned policy to collect data, which is both time-consuming and labor-intensive, especially for complex and risky tasks. In contrast, our MARG controller can be trained simultaneously with the policy. This concurrent training approach allows MARG to encounter a broader range of scenarios compared to offline data collection methods.
\begin{table}[H]
\centering
\caption{Comparison of Average Squared Estimated Linear Velocity Error across different Algorithms in Gap Terrains (3 Independent Seeds, 700 Time Steps).}
\label{average_error}
\renewcommand{\arraystretch}{1.1} 
\begin{tabular}{c|ccccc} 
\hline
\multirow{2}{*}{\textbf{Steps}} & \multicolumn{3}{c}{\textbf{Algorithms}} \\ 
\cline{2-4} 
& MorAL & MARG & DreamWaQ \\ 
\hline
\textbf{100}  & 0.326 $\pm$ 0.043 & \textbf{0.219 $\pm$ 0.021} & 0.330 $\pm$ 0.078 \\ 
\textbf{300}  & 0.311 $\pm$ 0.026 & \textbf{0.225 $\pm$ 0.029} & 0.294 $\pm$ 0.027 \\ 
\textbf{500}  & 0.301 $\pm$ 0.068 & \textbf{0.228 $\pm$ 0.031} & 0.308 $\pm$ 0.017 \\ 
\textbf{700}  & 0.270 $\pm$ 0.074 & \textbf{0.229 $\pm$ 0.029} & 0.304 $\pm$ 0.012 \\ 
\hline
\end{tabular}
\end{table}

For a fair comparison, we also compare average squared linear velocity errors during gap traversal, as shown in Table \ref{average_error}. Notably, the MARG algorithm consistently demonstrates the smallest absolute error and the lowest standard deviation across all tested steps. This performance highlights that MARG’s estimator net achieves exceptional precision and robustness in gap terrain scenarios.

\begin{figure*}[h]
   \centering
   \includegraphics[width=1.0\textwidth]{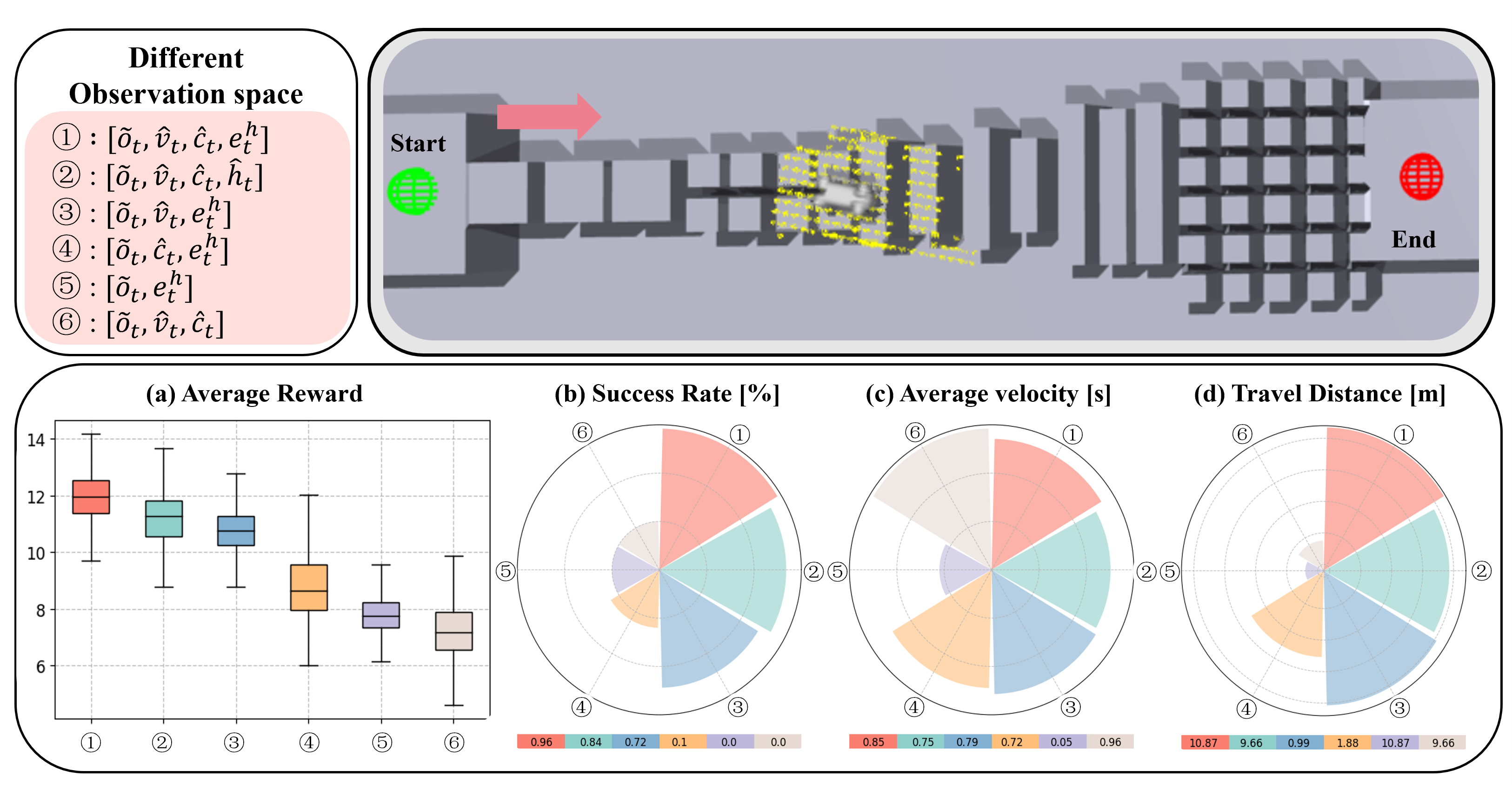}
   \caption{Evaluating the impact of 6 different observation spaces on the performance of the learned policy in risky terrains. The evaluation is based on four metrics: (a) average reward, (b) success rate, (c) average velocity, and (d) travel distance. Different combinations of sensor inputs in these observation spaces show different effects on the robot's performance in traversing discontinuous terrains.}
   \label{pics:compare_ablation}
\end{figure*}

\begin{figure*}[h]
   \centering
\includegraphics[width=1.0\textwidth, trim=0 0 0 0,clip]{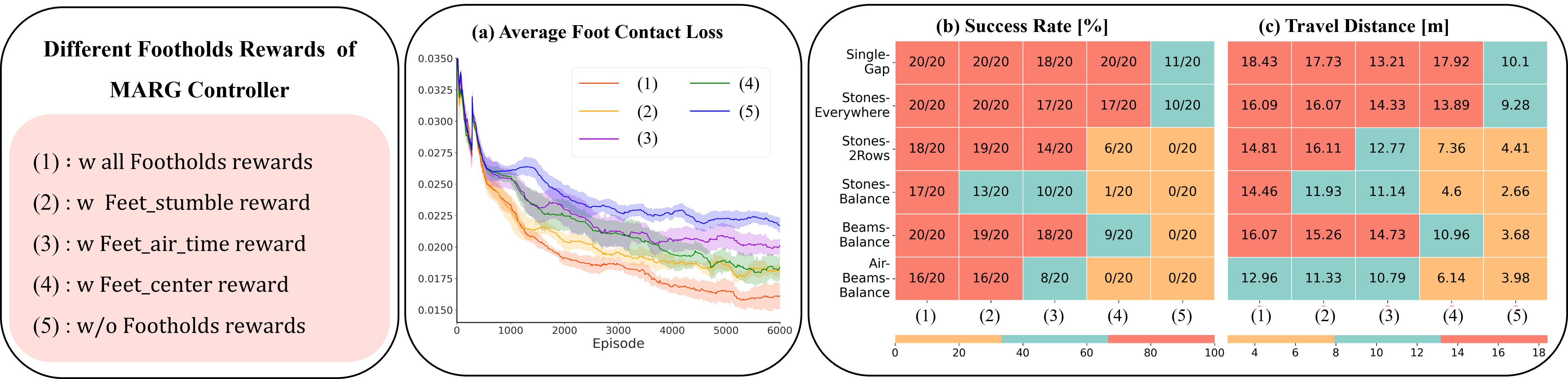}
   \caption{Evaluating the impact of different foothold rewards on the performance of the MARG in risky terrains. (a) represents the average foot contact loss of 5 different MARG controllers during training 4096 Unitree Go1 robots for 6000 episodes. Each curve’s shaded region represents the standard deviation of loss values across three different random seeds, indicating the uncertainty in the results. (b-c) represent the success rate and average traversal distance of five different MARG controllers when traversing six risky terrains. Each controller is evaluated through 20 repeat traversals of the level-6 risky terrains.}
   \label{feet_loss}
\end{figure*}

\subsection{Ablation Studies of Observation Space}

Fig. \ref{pics:compare_ablation} provides a comprehensive analysis of the performance of six different observation spaces on a risky terrain. We categorize observation spaces into six distinct types, each incorporating different sensor inputs, and evaluate these policies through four metrics: average reward, success rate, average velocity, and travel distance. 

Fig. \ref{pics:compare_ablation} (a) illustrates the reward distribution after the convergence of all controllers, specifically between episodes 4000 and 6000. The reward values decrease progressively from observation spaces $\textcircled{\scriptsize{1}}$ to $\textcircled{\scriptsize{6}}$, with $\textcircled{\scriptsize{1}}$ and $\textcircled{\scriptsize{2}}$ achieving the highest rewards and minimal variability. 

The robot is commanded to traverse the discontinuous terrains (including gaps, stones, and balance beams) at a speed of 0.9 m/s to further evaluate the different observation space combinations. After running the trials 50 times, we calculate the success rate, average velocity, and average travel distance, as shown in Figs. \ref{pics:compare_ablation} (b)-(d). Notably, $\textcircled{\scriptsize{1}}$, $\textcircled{\scriptsize{2}}$, and $\textcircled{\scriptsize{3}}$ exhibit higher success rates, average velocity, and long travel distance, suggesting that these combinations are effective for this risky task. The nearly zero success rates and significantly shorter travel distances observed for cases for $\textcircled{\scriptsize{4}}$, $\textcircled{\scriptsize{5}}$, and $\textcircled{\scriptsize{6}}$ demonstrated that without the body velocity or terrain map, completing the task is impossible. Due to the lack of body velocity, the average speed of $\textcircled{\scriptsize{5}}$ differs significantly from the command, resulting in stagnation because the robot perceives the gap terrain as a dangerous area. In addition, when the terrain map is lacking, $\textcircled{\scriptsize{6}}$ exhibits a high average velocity, often resulting in direct collisions with the terrain or falling into gaps. 
\begin{figure*}[h]
\centering
\includegraphics[width=0.99\textwidth , trim=10 5 5 15,clip]{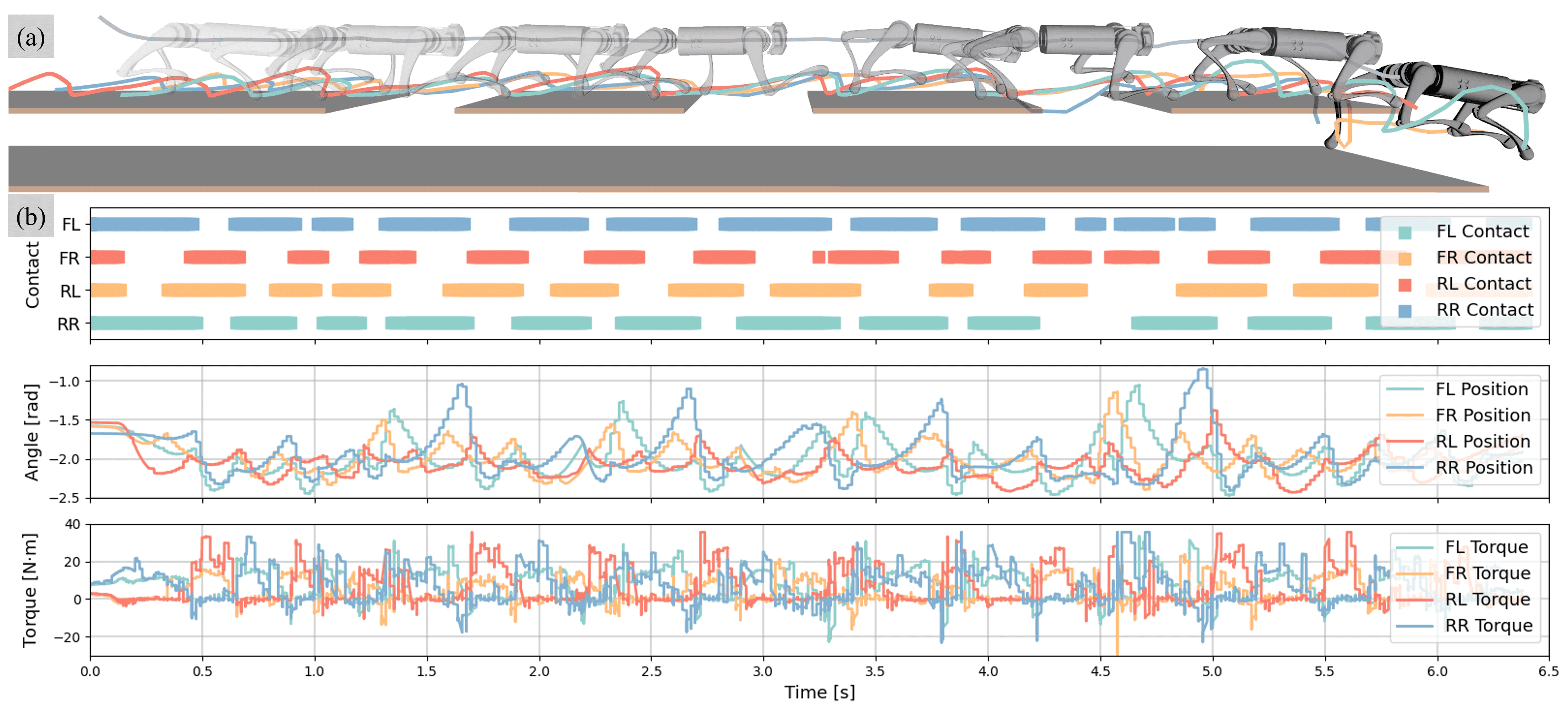}
   \caption{Dynamics analysis of quadruped robot traversing discontinuous gaps in gazebo simulation. (a) Snapshots of the simulated consecutive traversals across 40 cm gaps with body and foot positions in the world frame. (b) Variation curves of contact, joint angles, and torques during gap crossings. }
   \label{pics:dynamics_analysis}
\end{figure*}

Overall, we can conclude that different combinations of observation spaces significantly impact the performance of the learned policy, with both privileged information and exteroception terrain contributing to the performance. The terrain map is the most crucial factor, followed by the body velocity and the contact state. Additionally, using Elevation Net to extract elevation features $\boldsymbol{e}^h_t$ proves more efficient and stable than inputting the relative height map $\boldsymbol{\hat{h}}_t$.

\subsection{Ablation Studies of Footholds Rewards}
We also carry out detailed ablation experiments and comprehensively analyze the performance of five different MARG controllers with various foothold reward configurations in risky terrains. These controllers are classified according to five distinct footholds rewards setups: (1) MARG with all Footholds rewards; (2) MARG with Feet stumble reward; (3) MARG with Feet air time reward; (4) MARG with Feet center reward; (5) MARG without Footholds rewards. Fig. \ref{feet_loss} shows the three evaluation metrics: (a) average foot contact loss, (b) success rate, and (c) travel distance.

Fig. \ref{feet_loss} (a) represents the variation of the average foot contact loss, measured as $MSE(\hat{\boldsymbol{c}}_t, \boldsymbol{c}_t)$, of the five different MARG controllers during 6000 training episodes on risky terrains. Each curve corresponds to a distinct foothold reward configuration for the MARG controllers. As the number of training episodes increases, all curves exhibit a downward trend, signifying a gradual enhancement in the model's performance. Among them, (1), which is equipped with all foothold rewards, showcases optimal performance. Its loss value consistently stays at the lowest level and drops most swiftly. Conversely, (5), lacking foothold rewards, demonstrates the poorest performance, characterized by the highest loss value and a notably slower rate of decline. The performances of the remaining controllers fall between these two extremes. In conclusion, controllers with all foothold rewards can significantly enhance the accuracy of the Estimate network. Although single-item rewards (2) to (4) can improve prediction accuracy to some extent, none of them can match the performance of controllers equipped with all foothold rewards.
\begin{figure*}[h]
   \centering \includegraphics[width=0.99\textwidth, trim=5 5 5 5,clip]{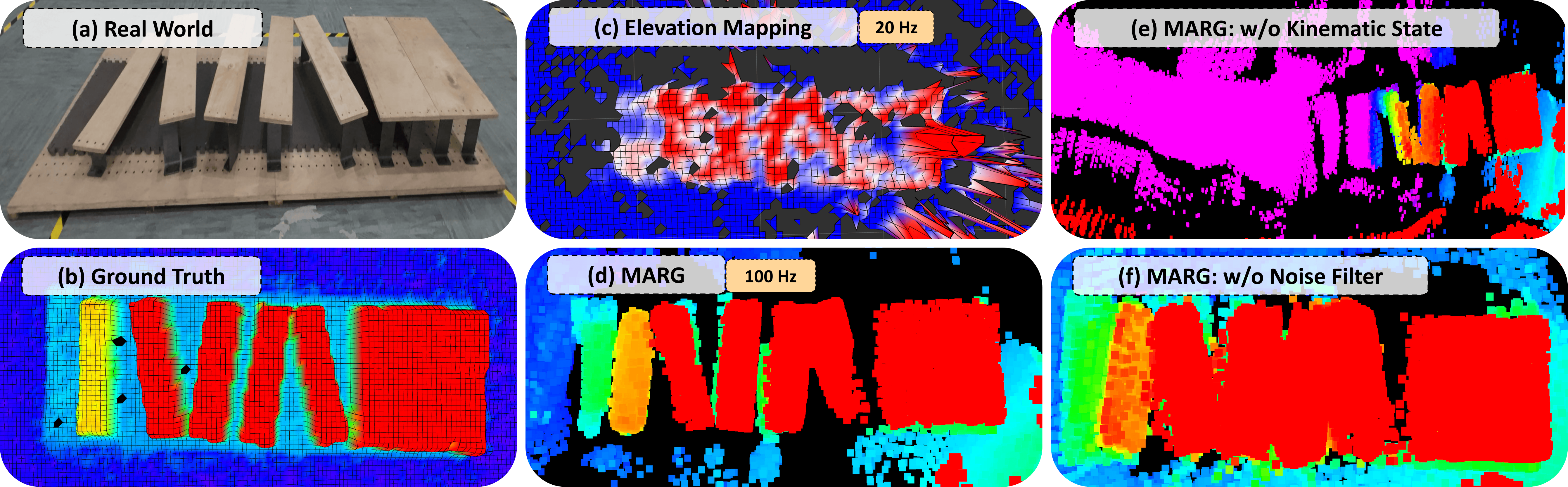}
   \caption{Comparison of two mapping methods with the ground truth. (a) shows the real-world snapshots, (b) is the ground truth for the mapping accuracy benchmark. The ground-truth terrain map is generated through handheld LiDAR scanning (Mid360) with FAST-LIO2 SLAM reconstruction (0.01 m resolution) and CloudCompare post-processing. (c) is an elevation mapping method with a 20 Hz update rate, showing less detail and more noise. (d) is the MARG method with a 100 Hz update rate, providing a clearer and more detailed representation. (e) shows the MARG method without kinematic state, and (f) shows the MARG method without noise filter, which highlights the importance of these components in mapping.}\label{pics:elv_msp}
\end{figure*}
Figs. \ref{feet_loss} (b) and (c) present the success rates and travel distance of five different MARG controllers when traversing through six distinct risky gap terrains. Each controller is evaluated via 20 repeated traversal tests on level-6 risky terrains. (1) shows significantly superior performance, outperforming the other controllers. In contrast, (5) exhibits the lowest success rate and travel distance. The success rates and traversal distances of (2) to (4) are intermediate, with values falling between those of (1) and (5). Among them, (2) shows the most outstanding performance, followed by (3), and (4).

Overall, the ablation experiments demonstrate the crucial role of foothold rewards in the performance of MARG. The controller, equipped with all foothold rewards, demonstrates a superior performance, outshining its counterparts across various evaluation metrics. In contrast, controllers lacking foothold rewards demonstrate significantly degraded performance, highlighting the indispensable role of these rewards in traversing risky gap terrains.

Different reward functions have varying degrees of influence on the performance of the controller. Among them, the foot stumble reward has proven to be the most effective, followed by the foot air time reward, with the foot center reward having the least impact. The feet stumble reward plays a crucial role, as it punishes the robot for tripping and encourages the controller to maintain a more stable gait, which is essential for successfully crossing risky terrain. The feet air time reward may contribute to optimizing the timing and coordination of leg movements, thereby ensuring efficient and safe traversal of uneven gap terrains. Although the impact of the foot center reward is comparatively modest, it can still play a role in preventing the robot from slipping off by influencing the foothold of the quadruped. By guiding the foothold of the feet, this reward mechanism helps avoid precarious positioning on the edges of gaps, thereby reducing the likelihood of missteps and subsequent falls.

\subsection{Dynamics analysis over risky terrains}
Fig. \ref{pics:dynamics_analysis} provides a comprehensive analysis of a quadruped robot's locomotion dynamics during the deployment of its learned policy in the Gazebo simulation. Fig. \ref{pics:dynamics_analysis} (a) offers a side view of the robot's trajectory and gait sequence, with colored lines representing the paths of individual legs and body, highlighting the spatial adjustments for balance and progression. (b) comprises three subplots: the first displays the contact sequence of each leg with the ground and the specific gait (i.e. trot), both of which are crucial for stability and propulsion; the second shows the angular positions of the robot's joints over time, indicating adaptive responses; and the third depicts the torque applied to each joint, reflecting active adjustments for balance and control. Our controller demonstrates the intricate coordination and control mechanisms required for the robot to achieve efficient and stable locomotion across risky gaps.

\begin{table}[H]
\caption{Accuracy evaluation on four different datasets}
\centering
\setlength\tabcolsep{4 pt}  
\renewcommand{\arraystretch}{1} 
\begin{threeparttable}
\begin{tabular}{c|cc|cc|cc|cc}
\bottomrule
\multirow{2}{*}{Method} & \multicolumn{2}{c|}{Park (m)} & \multicolumn{2}{c|}{Running (m)} & \multicolumn{2}{c|}{Indoor (m)} & \multicolumn{2}{c}{Corridor (m)} \\
\cline{2-3} \cline{4-5} \cline{6-7} \cline{8-9}
 & APE& RPE & APE & RPE & APE  & RPE& APE& RPE\\
\hline
EKF & 4.44 & 0.21 & 0.64 & 0.19 & 0.15 & 0.10 & 2.10 & 0.17 \\
FAST-LIO2 & 0.30 & 0.15 & 0.30 & 0.18 & 0.19 & 0.25 & 0.49 & 0.19 \\
LIO-SAM & 0.21 & 0.15 & 0.10 & 0.17 & 1.24 & 0.71 & 1.04\tnote{1} & 0.23\tnote{1}\\
A-LOAM & 2.86 & 0.19 & 0.09 & 0.15 & 0.06 & 0.08 & 0.51 & \textbf{0.14} \\
MARG (Ours) & \textbf{0.19} & \textbf{0.15} & \textbf{0.04} & \textbf{0.04} & \textbf{0.05} & \textbf{0.06} & \textbf{0.40} & 0.21 \\
\bottomrule
\end{tabular}
\begin{tablenotes}
    \item [1] This method fails, and we cut off the drift part.
\end{tablenotes}
\label{tab:APE_RPE}
\end{threeparttable}
\end{table}

\subsection{Analysis of the TMG model}
Table. \ref{tab:APE_RPE} presents an accuracy evaluation of different localization methods, including MARG, EKF~\cite{bledt2018cheetah}, LIO-SAM~\cite{shan2020lio}, A-LOAM~\cite{zhang2014loam}, FAST-LIO2~\cite{xu2022fast} and Ground Truth, based on Absolute Pose Error (APE) and Relative Pose Error (RPE) across a dataset \cite{ou2024leg} comprising Park, Running, Indoor, and Corridor scenarios. MARG consistently achieves the lowest APE and RPE across most environments, highlighting its precision and reliability. 

\begin{figure}[h]
   \centering
   \includegraphics[width=0.48\textwidth, trim=2 2 2 2,clip]{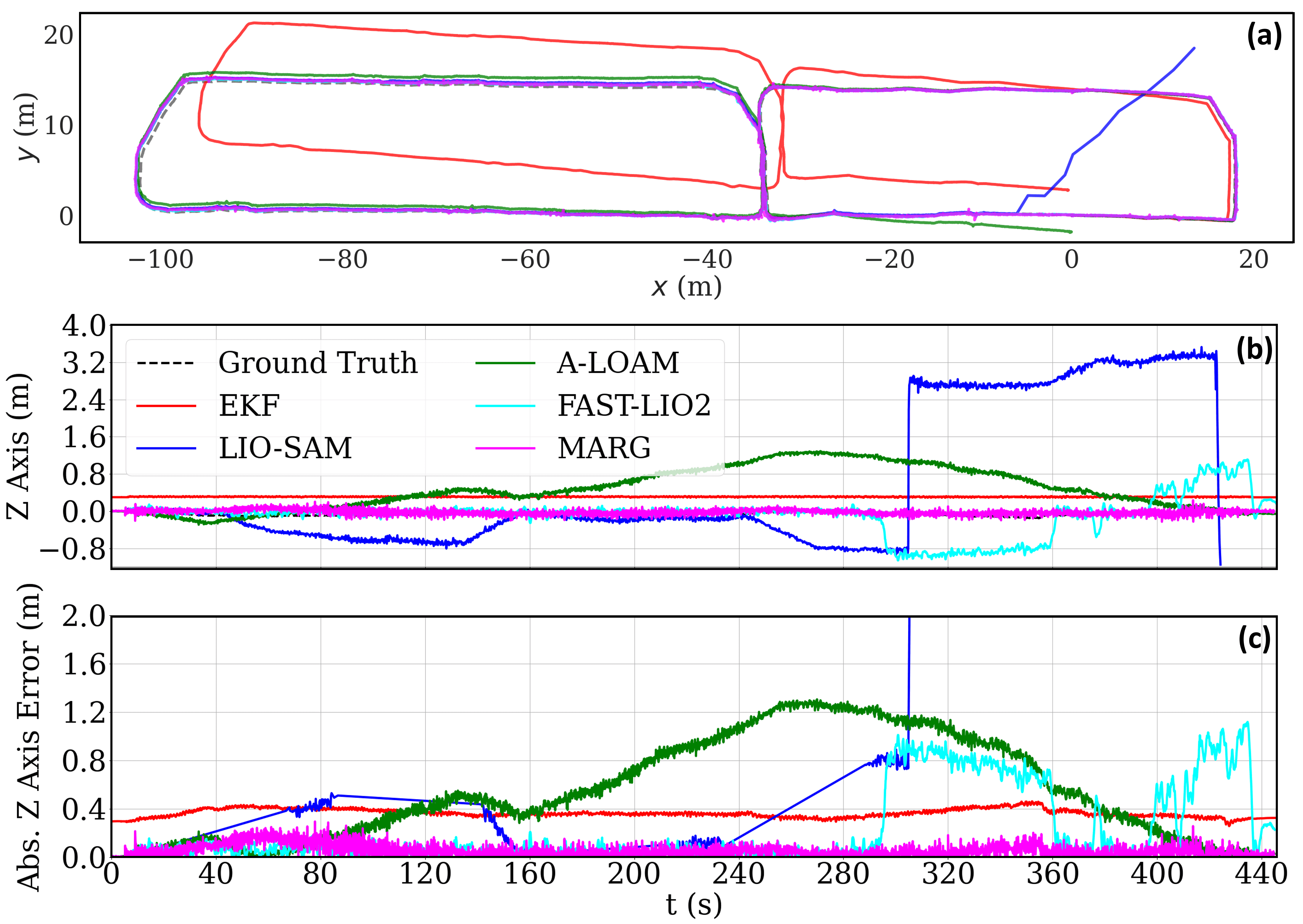}
   \caption{Compare the five localization algorithms on the corridor dataset in terms of the long-distance trajectory (a), z-axis position (b), and absolute z-axis error (c).}
   \label{pics:compare_TMG}
\end{figure}

\begin{figure*}[t]
\centerline{\includegraphics[width=0.98\linewidth,trim=10 15 15 5, clip]{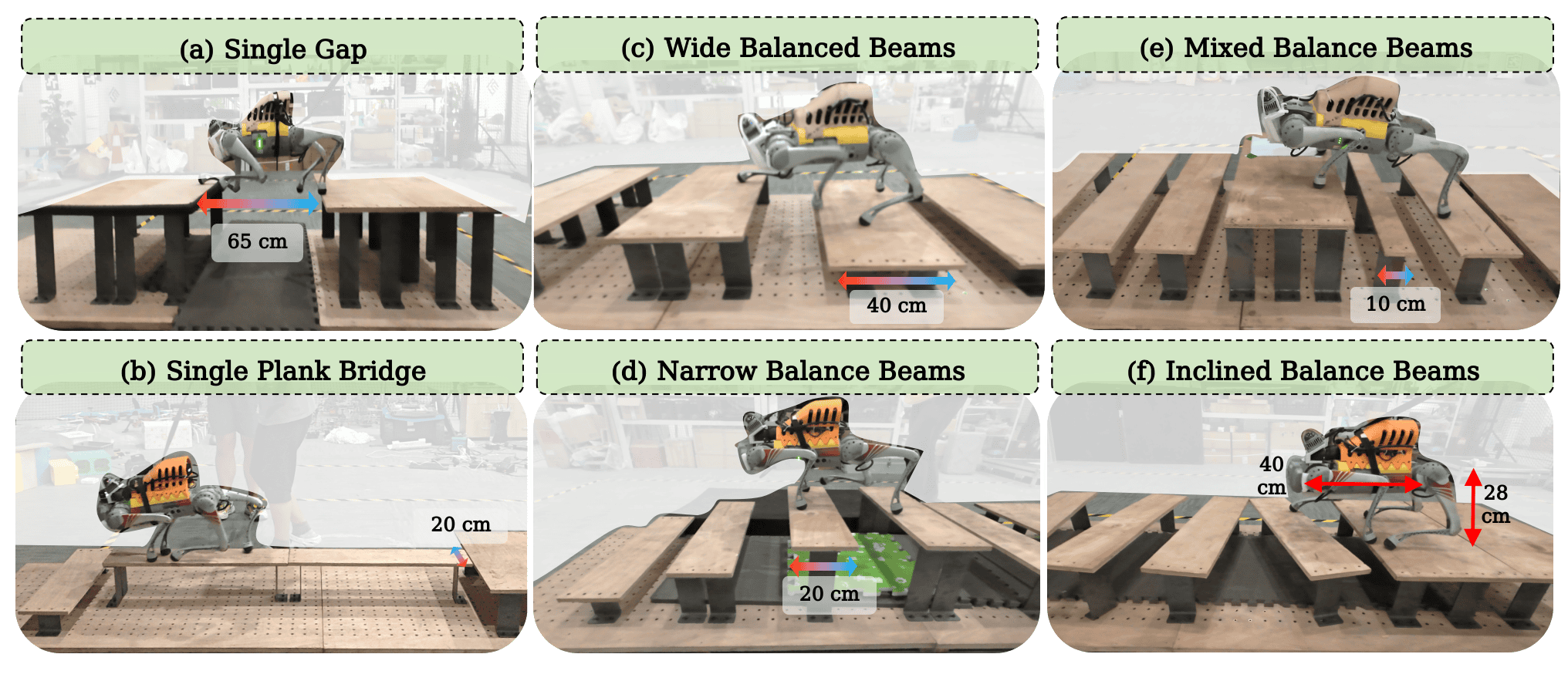}}
    \caption{Deployment scenarios on the quadrupedal robot Go1 under various risky gaps: (a) traversing a 65 cm wide gap using a trot gait. (b) walking on a 20 cm single-plank bridge to demonstrate balance ability. (c\text{-}f) crossing balance beams with varying heights, widths, and inclination angles.}
   \label{pics:deployment_real}
\end{figure*}
\begin{figure}[t]
   \centering
   \includegraphics[width=0.49\textwidth, height=0.3\textwidth,  trim=3 3 10 3,clip]{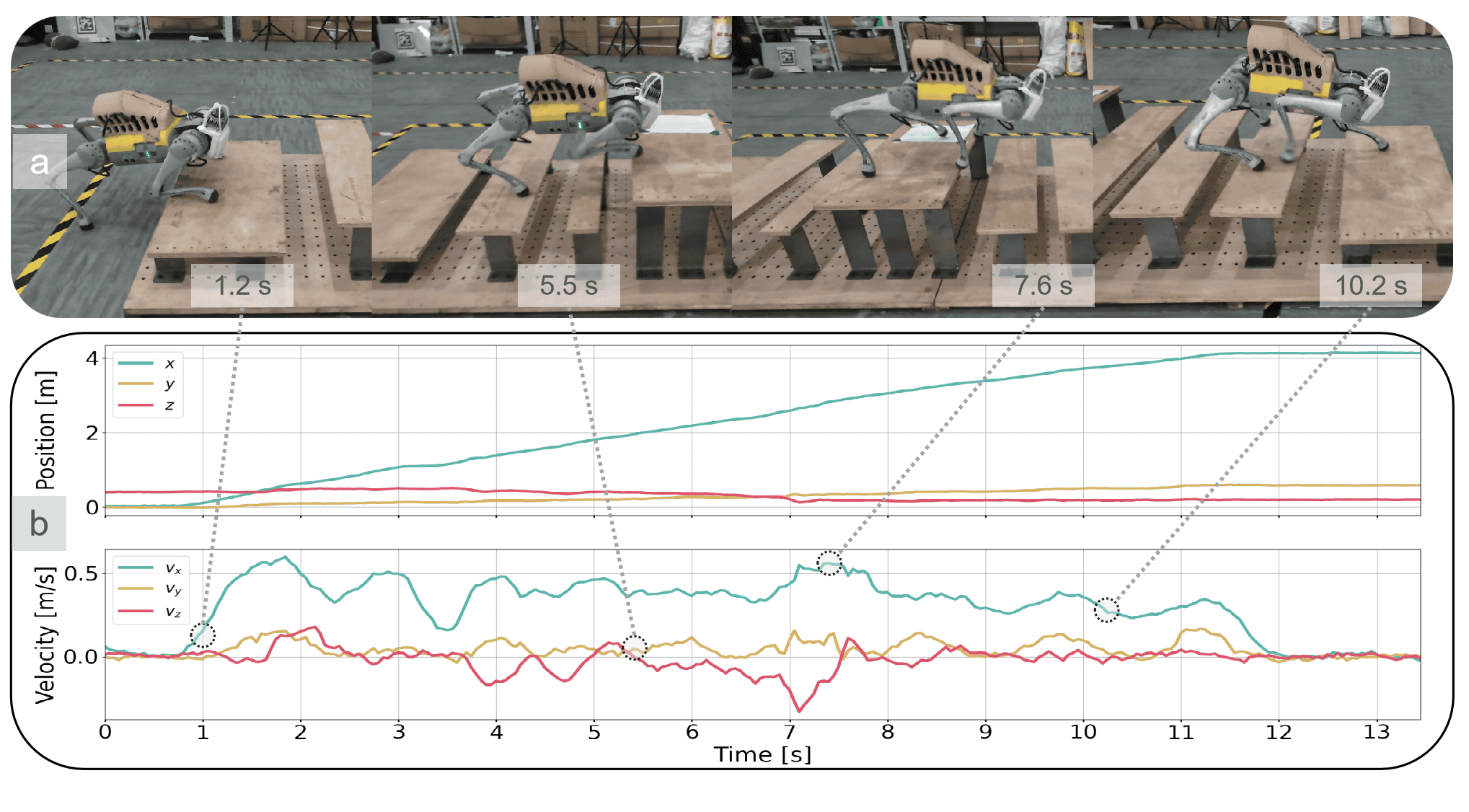}
    \caption{Learned motions on balance beams in real-world experiments. (a) presents the motion snapshots of the robot during traversal in real-world scenarios, offering a visual record of its movement process. (b) exhibits the position and velocity of the robot, which shows the kinematic characteristics during the motion.}
   \label{pics:compare_real_dynamic}
\end{figure}
Meanwhile, we also compare the performance of these five localization algorithms in long-distance trajectory drift in these datasets, as shown in Fig. \ref{pics:compare_TMG} (a).  MARG and FAST-LIO2 closely follow this path, indicating high accuracy, while EKF shows significant deviation, particularly in turns. LIO-SAM deviates after the path turns, and A-LOAM shows some inconsistencies. For the 3D elevation tracking shown in Fig. \ref{pics:compare_TMG} (b), the ground truth line remains stable, with MARG aligning closely, demonstrating robust performance. In contrast, LIO-SAM exhibits significant fluctuations, indicating poor elevation accuracy. A-LOAM and FAST-LIO2 have deviations, as can be seen in Fig. \ref{pics:compare_TMG} (c). These deviations cause errors in the elevation map, which can harm quadrupedal locomotion. The fact that only our MARG maintains an accurate height estimation highlights the effectiveness of our proposed method.
Overall, MARG excels in both path and elevation tracking, highlighting the importance of choosing a reliable localization algorithm in complex environments.

\subsection{Real-world transfer on indoor risky gap terrains}
Fig. \ref{pics:elv_msp} compares two mapping methods for accurately representing a real-world structure. The top quadrants (a-b) depict the actual physical setup and the ground truth, serving as the benchmark for mapping accuracy. 
The elevation mapping method \cite{fankhauser2018probabilistic} operates at a slower update rate of 20 Hz, as illustrated in Fig. \ref{pics:elv_msp} (c). As a result, it captures less detail and exhibits more noise, highlighting its limitations in accurately rendering fine structural features. In contrast, the proposed MARG method (d), updating at 100 Hz, provides a significantly clearer and more detailed representation that closely aligns with the ground truth. In the absence of the kinematic state, as shown in (e), the elevation map exhibits significant deviations due to the IMU becoming oversaturated when the robot experiences impacts during movement. Compared to (f), it is evident that the noise filter effectively eliminates uncertainties caused by the beam angle and manages the beam divergence characteristics of the LiDAR sensor. These comparisons underscore MARG's superior capability in achieving high-resolution and accurate mapping at higher frequencies, thus demonstrating its effectiveness over lower-frequency methods.

Extensive tests have been carried out on various risky terrains, as shown in Fig. \ref{pics:deployment_real} and the supplementary video.

\begin{figure*}[h]
   \centering
   \subfigure[These outdoor experiments are conducted on diverse terrains, including gardens, slopes, gaps, stairs, and so on. The tests are performed under varying illumination conditions, ranging from bright to dark environments. These experiments systematically evaluate the performance of MARG in different real-world terrains.]
   {\includegraphics[width=0.96\textwidth, trim=1 1 1 1,clip]{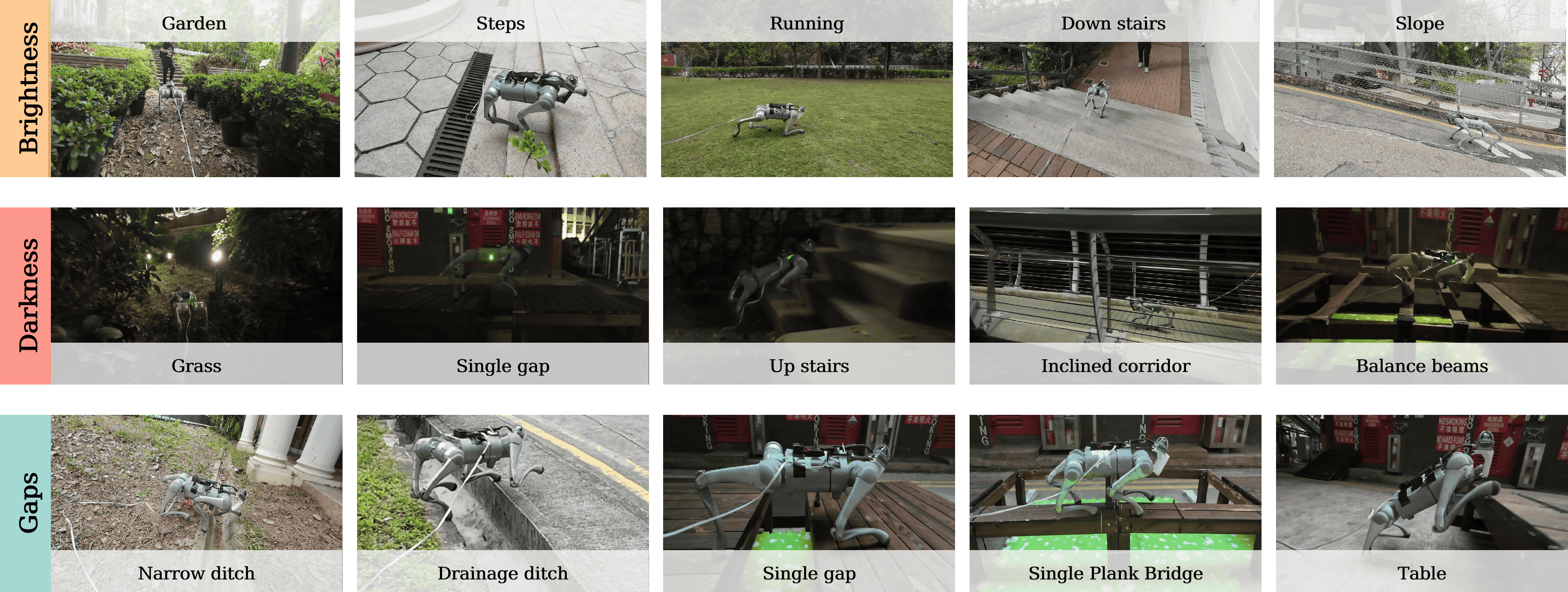}} 
   \subfigure[The experimental route covers seven key terrain nodes from A to G, including various obstacles such as steps, platforms, and guardrails. Meanwhile, through real-scene photos and photos of elevation points restored by TMG, the motion control performance is verified from multiple dimensions, comprehensively evaluating the stability and reliability of the controller in complex outdoor terrains.]{
    \includegraphics[width=0.96\textwidth, trim=1 1 1 1,clip]{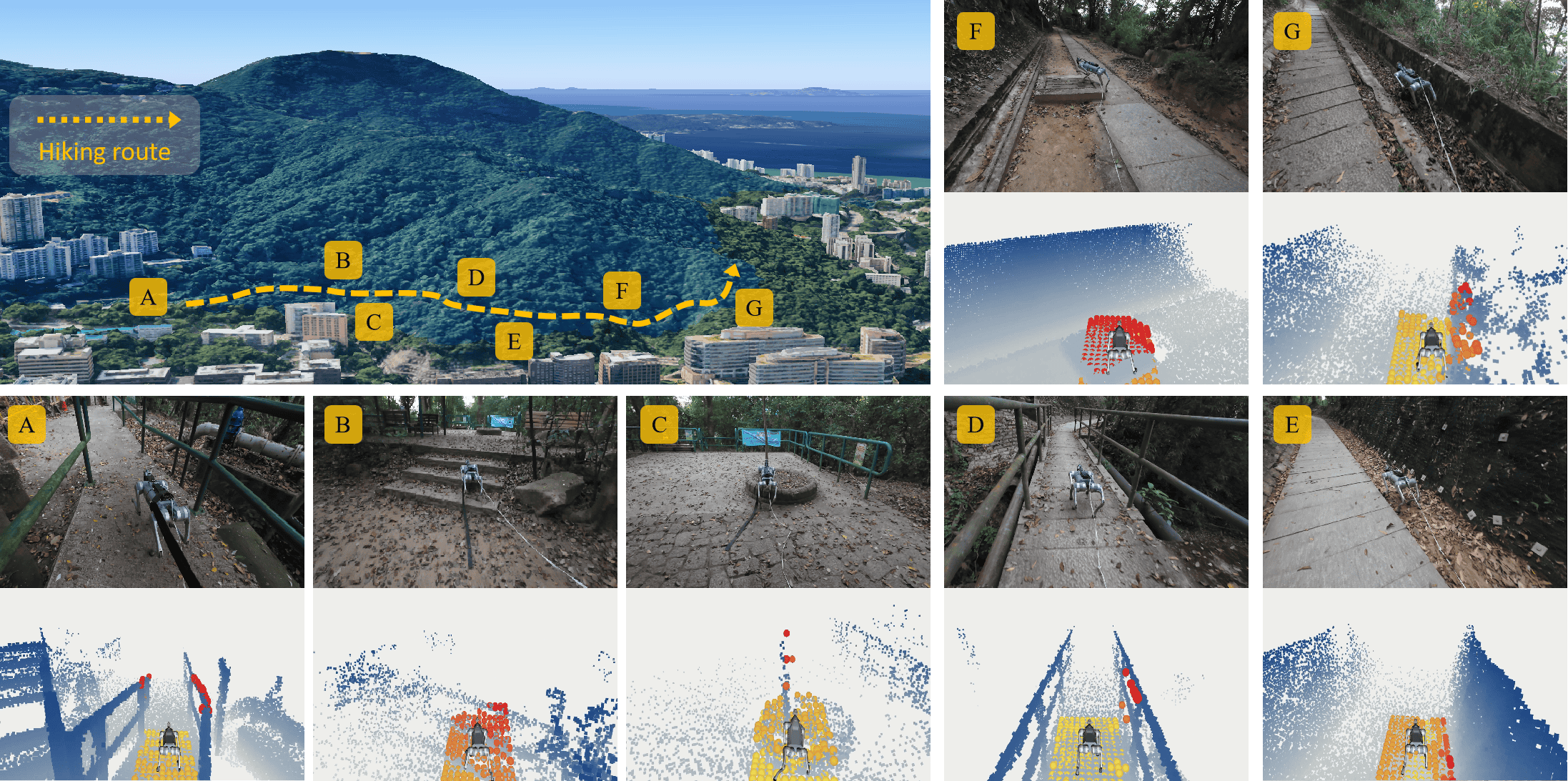}}
   \caption{Robustness testing of the Unitree Go2 robot in diverse challenging terrains: (a) Outdoor experiments are conducted on the HKU campus, (b) Robustness tests are carried out on the hiking route in Lung Fu Shan Country Park.}\label{new_outdoors}
\end{figure*}
\subsubsection{Large gaps}
Our controller can cross large gaps of 65 cm (1.6 $\times$ robot length) with a trot gait, as shown in Fig. \ref{pics:deployment_real} (a). The MARG guides the robot to adjust the leg extension angle and the magnitude of force to maintain a stable posture and a safe landing point when crossing the gap. 

\subsubsection{Single-plank bridge}
In addition, the controller can cross a 20 cm single-plank bridge scene (0.7 $\times$ robot width), demonstrating the robot's balance ability on narrow surfaces, as shown in Fig. \ref{pics:deployment_real} (b). The MARG controller can quickly respond to any tiny shaking or imbalance and dynamically adjust the support position and strength of the legs to ensure that the robot always maintains a stable center of gravity and avoids slipping or overturning. 

\subsubsection{Beams}
Furthermore, we have conducted experiments on balance beams with different heights, widths, and inclined angles. The gaps between beams constantly change, as shown in Fig. \ref{pics:deployment_real} (c-f). These complex and variable experiments pose a severe challenge to the locomotion controller. For example, beams of different heights require the robot to precisely adjust the extension degree of its legs to adapt to different ascending and descending slopes. Various widths of breams and the constantly changing gap further increase the difficulty of the robot locomotion. The controller needs to accurately plan the footholds and leg movements at each step to ensure the robot can pass through these risky terrains stably and safely. Beams with different inclined angles (approximately 10 to 15 degrees) also demonstrate the controller's ability to perceive terrain and adjust the robot. These experiments illustrate the effectiveness and adaptability of the MARG controller in practical applications and highlight its significant advantages in solving the problem of quadruped robots walking on dangerous terrains.

\subsubsection{Mixed risky terrain}
Fig. \ref{pics:compare_real_dynamic} illustrates the quadruped robot locomotion across a series of uneven gaps, demonstrating its ability to maintain balance and stability. The top sequence shows the robot's progression over time, while the bottom graphs depict its position and velocity dynamics. The data highlights the robot's effective control mechanisms, enabling it to adapt its movement and maintain consistent velocity, even when encountering complex terrain features. 

\subsection{Real-world transfer on outdoor terrains}
Fig. \ref{new_outdoors} (a) shows the comprehensive outdoor experimental scenarios conducted on the campus of the University of Hong Kong. We conduct systematic tests on Unitree Go2 to comprehensively evaluate the performance of MARG on diverse outdoor risky terrains, including 9 cm beams, 18 cm single-plank bridges, and various sizes of gaps. Beyond these risky terrains, the robot is also commanded to traverse various other conventional terrains, such as
gardens, slopes, gaps, and stairs. Additionally, the environment tests are conducted on a wide range of lighting conditions, from bright daylight to dark. 
The results clearly indicate that the TMG model empowers robots to execute tasks efficiently across diverse campus outdoor environments, thereby underscoring its remarkable adaptability and stability.

Fig. \ref{new_outdoors} (b)  highlights the robustness test of the hiking route in Lung Fu Shan Country Park. The yellow dashed line marks the experimental route, and seven key terrain nodes from A to G are selected. These terrains include various complex obstacles such as steps, platforms, and guardrails, comprehensively simulating the typical challenging scenarios in the natural outdoor environment. By presenting the real-scene photos and the submitted video, the entire movement process of the robot outdoors is intuitively demonstrated. 
Moreover, the generated terrain map clearly demonstrates its dynamic changes during the robot's movement, enabling an intuitive grasp of how the robot senses and adapts to diverse terrain elevations and contours. The experimental results show that the TMG model can operate stably and effectively deal with various challenges of the outdoor terrain.

\subsection{Limitations}\label{limitations}

\subsubsection{Dependency on sensor accuracy}
The performance of our system depends critically on the accuracy of the point cloud data captured by the LiDAR sensor. Although we have proposed a method to enhance point cloud fidelity, factors such as sensor noise, dust accumulation, calibration errors and the vertical field of view still influence the quality of the elevation maps which will shift or distort the point cloud, impair obstacle detection, and pose significant safety risks in high-precision robotic applications.
\subsubsection{Limited Gait Diversity}
MARG can only achieve the trot gait. Previous studies \cite{alexander2003principles}, \cite{margolis2023walk} have shown that quadruped animals can adapt to different terrains through various gaits. 
In contrast, our current approach has not fully explored the locomotion potential of the robot. 


\section{CONCLUSIONS}\label{conclutions}
This study successfully demonstrated the excellent locomotion performance of the MARG controller for quadruped robots. The MARG controller accurately predicted the body velocity and the contact state of each foot, ensuring that the robot adjusted its posture timely manner under imbalances. Meanwhile, the three foot-related rewards proved to be extremely effective in guiding the robot to explore safe footholds. In addition, the TMG model relied solely on a single LiDAR to generate accurate terrain maps, simplifying the hardware deployment. Thus, the policy trained in the simulation could be directly transferred to the real world, significantly enhancing the adaptability and practicality of the robots. The experiments demonstrated that the MARG controller was stable and effective across risky tasks, successfully balancing safety, stability, and efficiency during locomotion. In the future, we will further explore ways to optimize the controller's performance and expand its application to a broader range of scenarios, including soft, slippery, and unstable risky terrains.

\bibliographystyle{IEEEtran}
\bibliography{root}


\begin{IEEEbiography}[{\includegraphics[width=2in,height=1.25in,clip,keepaspectratio]{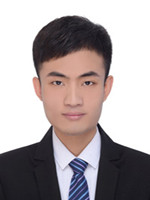}}]
{Yinzhao Dong} received the B.Sc. degree from Jilin University, Changchun, China, and the M.Sc. degree from Dalian University of Technology, Dalian, China, in 2018 and 2021, respectively. He is currently working toward the Ph.D. degree in mechanical engineering with the Adaptive Robotic Controls Lab(ArcLab), the University of Hong Kong, Hong Kong. 

His research interests include legged robotics, motion planning, and reinforcement learning.
\end{IEEEbiography}

\begin{IEEEbiography}[{\includegraphics[width=1in,height=1.25in,clip,keepaspectratio]{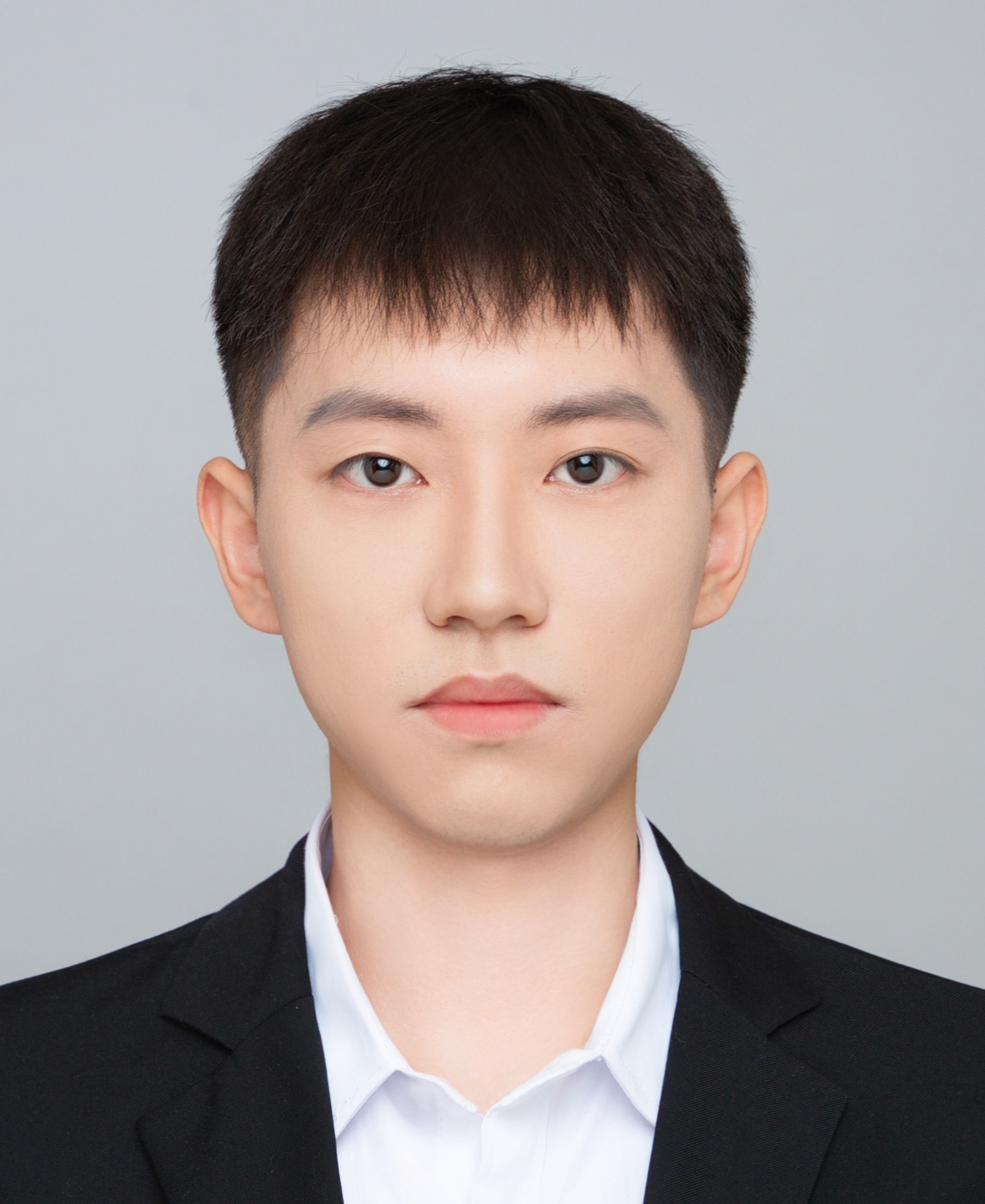}}]
{Ji Ma} received the B.Sc. degree in Information Engineering from Jilin University, Changchun, China, in 2023. He is currently working toward the Ph.D. degree in mechanical engineering with the Adaptive Robotic Controls Lab(ArcLab), the University of Hong Kong, Hong Kong. 

His research interests include legged robotics, motion planning, and reinforcement learning.
\end{IEEEbiography}

\begin{IEEEbiography}[{\includegraphics[width=1in,height=1.25in,clip,keepaspectratio]{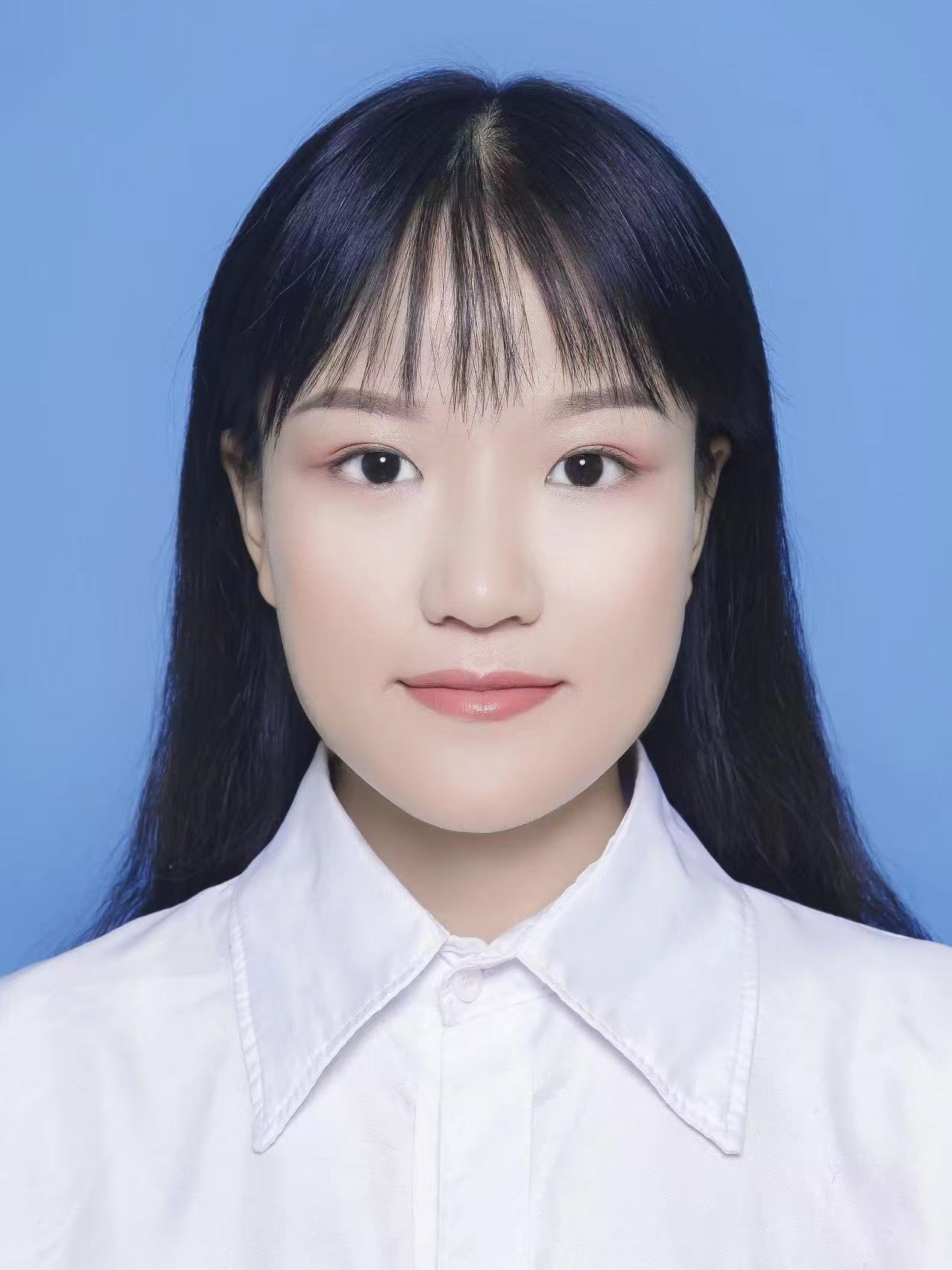}}]
{Liu Zhao} received the B.Sc. degree in Automation and M.Sc. degree in Control Science and Engineering both from Harbin Institute of Technology (HIT), in 2021 and 2023. She is currently working toward the Ph.D. degree in mechanical engineering with the Adaptive Robotic Controls Lab(ArcLab), the University of Hong Kong, Hong Kong. 

Her research interests include vision, navigation, and control of legged robotics.
\end{IEEEbiography}

\begin{IEEEbiography}[{\includegraphics[width=1in,height=1.25in,clip,keepaspectratio]{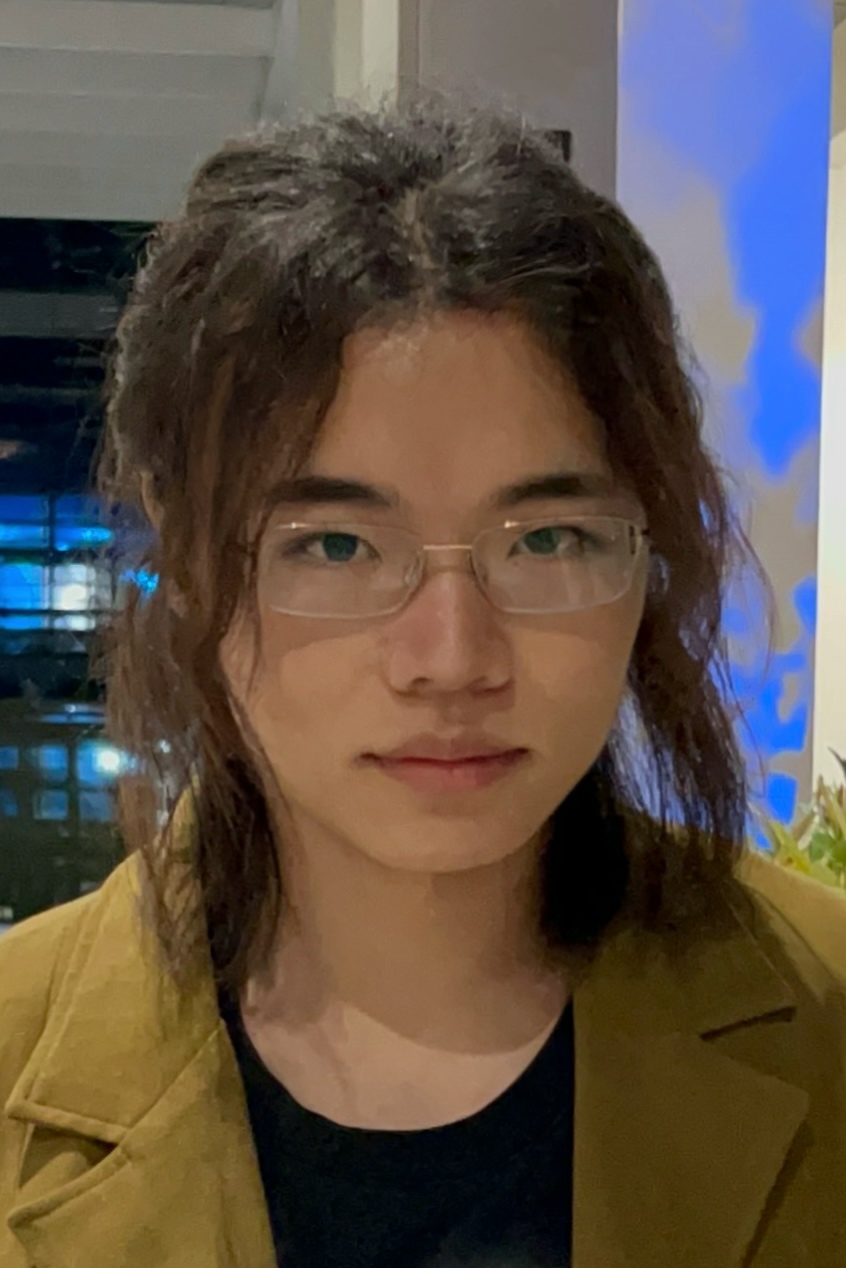}}]
{Wanyue Li}(Graduate Student Member, IEEE) received the B.Sc. degree in Computer Science and Technology from South China Agricultural University in 2019, and the M.Sc. degree in Artificial Intelligence from Sun Yat-sen University in 2023. He is now a Ph.D. candidate in mechanical engineering with the Adaptive Robotic Controls Lab (ArcLab), the University of Hong Kong, Hong Kong. 

His research interests include humanoid robot locomotion control, trajectory optimization, and reinforcement learning.
\end{IEEEbiography}

\begin{IEEEbiography}[{\includegraphics[width=1in,height=1.25in,clip,keepaspectratio]{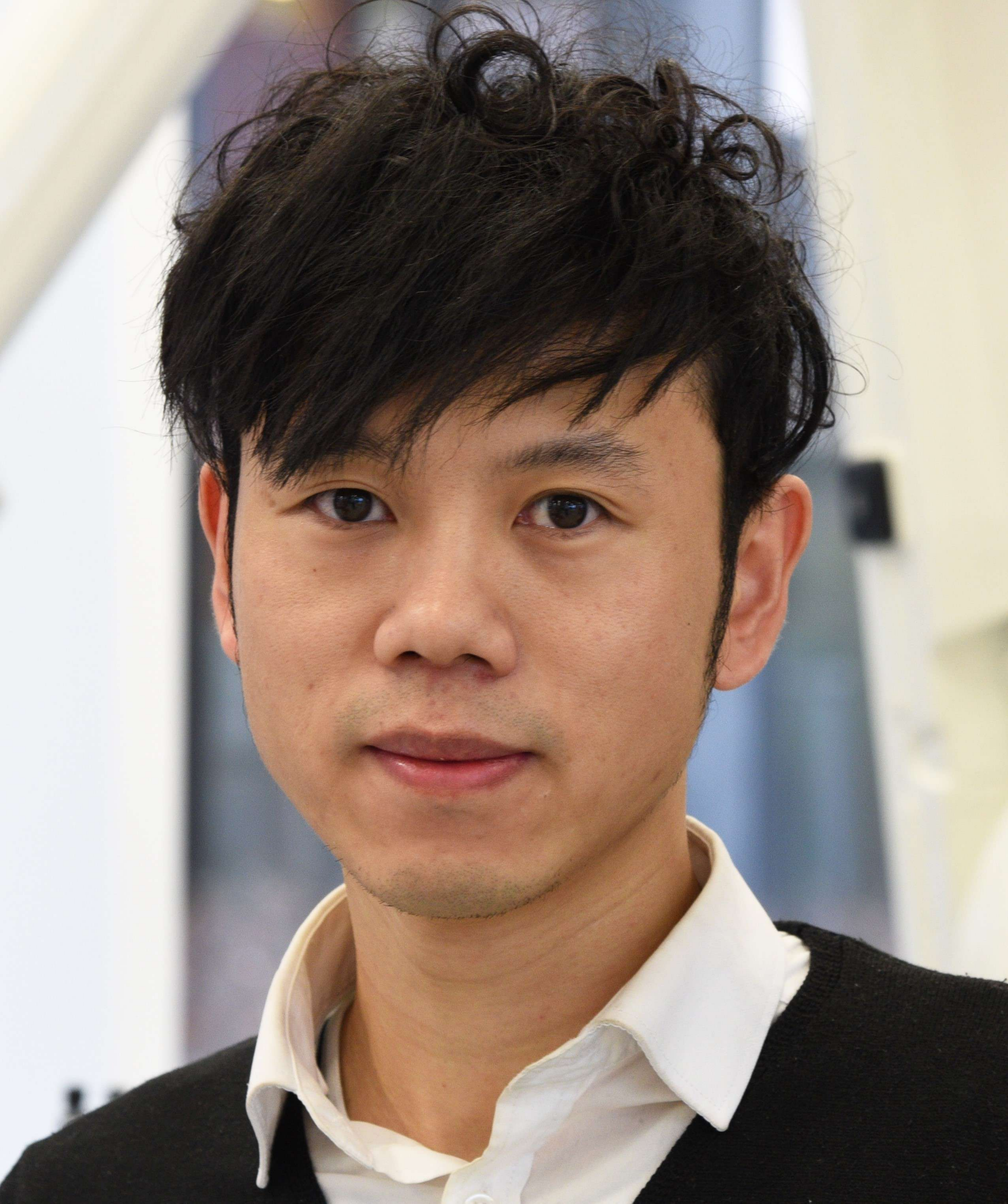}}]{Peng Lu}
obtained his BSc degree in automatic control and MSc degree in nonlinear flight control both from Northwestern Polytechnical University (NPU). He continued his journey on flight control at Delft University of Technology (TU Delft) where he received his PhD degree in 2016. After that, he shifted a bit from flight control and started to explore control for ground/construction robotics at ETH Zurich (ADRL lab) as a Postdoc researcher in 2016. He also had a short but nice journey at University of Zurich \& ETH Zurich (RPG group) where he was working on vision-based control for UAVs as a Postdoc researcher. He was an assistant professor in autonomous UAVs and robotics at Hong Kong Polytechnic University prior to joining the University of Hong Kong in 2020.

Prof. Lu has received several awards such as 3rd place in 2019 IROS autonomous drone racing competition and best graduate student paper finalist in AIAA GNC. He serves as an associate editor for IROS and session chair/co-chair for conferences like IROS and AIAA GNC for several times. He also gave a number of invited/keynote speeches at multiple conferences, universities and research institutes.
\end{IEEEbiography}

\end{document}